\documentclass{siamltex}

\usepackage{amsmath, amssymb, graphicx, url, algorithm2e}
\usepackage{bm}
\usepackage{color}
\usepackage{array}

\DeclareMathAccent{\maxvec}{\mathord}{letters}{"7E}
\DeclareMathOperator*{\argmin}{argmin}

\usepackage{pifont}
\newtheorem{remark}{Remark}[section]

\newcommand{\PBB}[1]{\textcolor{black}{#1}}

\newcommand{\eg}{\textit{e.\@{}g.\@{}}}
\newcommand{\ie}{\textit{i.\@{}e.\@{}}}
\newcommand{\etc}{\textit{etc.\@{}}}
\newcommand{\eetc}{\textit{etc}}

\newcommand{\vs}{\textit{vs.\@{}}}

\title{Development, Demonstration, and Validation of Data-driven Compact 
Diode Models for Circuit Simulation and Analysis}

\usepackage{times}
\author{
K. Aadithya\thanks{Radiation and electrical sciences, Center~1300, Sandia National Laboratories}
\and
P. Kuberry\thanks{Computing Research, Center~1400, Sandia National Laboratories}
\and
B. Paskaleva\footnotemark[1]
\and
P. Bochev\footnotemark[2]
\and 
K. Leeson\footnotemark[1]
\and 
A. Mar\thanks{Integrated Military Systems Development, Center~5400, Sandia National Laboratories}
\and
T. Mei\footnotemark[1]
\and
E. Keiter\footnotemark[1]
}

\begin{document}
\maketitle

\renewcommand{\baselinestretch}{0.98}
\selectfont

\begin{abstract}
Compact semiconductor device models are essential for efficiently designing and
analyzing large circuits. However, traditional compact model development
requires a large amount of manual effort and can span many years. Moreover,
inclusion of new physics (\eg{},~radiation effects) into an existing compact model is
not trivial and may require redevelopment from scratch. Machine Learning (ML)
techniques have the potential to automate and significantly speed up the
development of compact models. In addition, ML provides a range of modeling
options that can be used to develop hierarchies of compact models tailored to
specific circuit design stages. In this paper, we explore three such
options:~(1)~table-based interpolation, (2)~Generalized Moving Least-Squares,
and~(3)~feed-forward Deep Neural Networks, to develop compact models for a p-n
junction diode. We evaluate the performance of these ``data-driven'' compact
models by~(1)~comparing their voltage-current characteristics against laboratory
data, and~(2)~building a bridge rectifier circuit using these devices,
predicting the circuit's behavior using SPICE-like circuit simulations, and then
comparing these predictions against laboratory measurements of the same circuit.
\end{abstract}


\section{Introduction}
\label{sec:intro}

Circuit simulation, sometimes referred to as SPICE simulation, is foundational
to modern circuit design \cite{Nagel_75_THESIS}. In circuit simulation,
so-called ``compact models'' are used to capture the dynamics of voltages,
currents, and charges in individual circuit components (\eg{},~transistors,
diodes, resistors, capacitors,~\etc{}). Given a circuit composed of many such
components connected to each other, a circuit simulator combines the compact
models of the individual components to enforce Kirchhoff's voltage and current
laws across the network. This is done by building a non-linear system of
Differential-Algebraic Equations (DAEs); each equation in this system is of the
form $a+b+c+\ldots{} = 0$, where the individual terms ($a$, $b$, $c$,
$\ldots{}$) are provided by compact models. The circuit simulator numerically
solves the system of equations as a whole, using a combination of time-stepping
algorithms and non-linear
solvers~\cite{Keiter_19_TECHREPORT,Keiter_19a_TECHREPORT}.

As modern circuits can easily have many thousands of components (leading to DAE
systems of similar size), it is important that each individual compact model be
computationally inexpensive. In practice, typical compact models consist of only
a handful of algebraic and ordinary differential equations, which are generally
a combination of empirical formulas and simplified solutions to semiconductor
transport equations. 

\begin{remark}
\label{rem:TCAD}
In addition to compact models, there also exist ``first-principles'', or TCAD
(Technology Computer-Aided Design), semiconductor device models that typically
provide much more accurate descriptions of device physics over a wide range of
operating conditions. Such TCAD models work by predicting the electric field at
every point within a three-dimensional semiconductor device, and the resulting
movement of charge carriers (electrons and holes) in the device.  But doing so
is computationally very expensive. Therefore, TCAD codes such as
Charon~\cite{charon} are orders of magnitude slower than
compact models. For this reason, TCAD is almost never used directly in a circuit
simulator. Indeed, one can view compact models used in circuit simulators as
much faster reduced-order approximations of corresponding TCAD models.
\end{remark}

Developing compact models for new electrical components is a difficult task 
requiring extensive expertise in solid state physics, circuit design, model 
calibration, and numerical analysis. For example, the BSIM family of compact 
models for Metal-Oxide-Semiconductor (MOS) transistors is the result of over 20 
years of work by Prof.~Chenming Hu and his team of PhD students and postdocs at 
UC Berkeley \cite{BSIM1,BSIM2,BSIM3}. Besides long development times that can 
span many person-years, reliance on simplified solutions in traditional compact 
models may compromise their ability to generalize. As a result, adding new 
physics to a legacy compact model (\eg{}, to scale the model down to a more
advanced CMOS technology node, to take into account radiation effects in harsh 
environments,~\etc{}) often requires extensive redevelopment. 

We believe that a ``data-driven'' approach,~\ie{},~using Machine Learning (ML)
techniques, appropriately specialized for the semiconductor device physics
domain, to automate the development of compact models directly from electrical
data, has the potential to overcome the challenges above. Moreover, ML 
techniques provide a wide range of regression methods that can be used to 
develop hierarchies of compact models tailored to specific circuit design 
stages, specific circuit simulation tasks, and even specific compute 
infrastructures. Indeed, in a world where compact model development is fully 
automated, where a variety of compact models capturing different facets of a 
device's behaviour, with different computational efficiency and accuracy tradeoffs, can 
all be generated at the push of a button, it is conceivable that a circuit
designer would use a different compact model for initial exploration and a 
different one for late-stage design, one for timing analysis and another one 
for sensitivity analysis, one for CPU simulation and one for GPU simulation, 
and so on. The benefits would be immense~--~enabling rapid, cost-effective, and 
robust circuit design flows calibrated against real-world electrical data from 
day one. 

Thus, we believe that the application of ML techniques to compact model
development should be thoroughly and systematically explored. To that end, in
this paper, we investigate three markedly different ML regression
approaches~--~Table-Based Interpolation (TBI), Generalized Moving Least-Squares
(GMLS), and Deep Neural Networks (DNNs)~--~for developing data-driven compact
device models. Specifically, we apply these approaches to develop compact models
for a 1N4148 high-speed switching diode, a common mass-produced semiconductor
device with well-documented electrical and thermal characteristics
\cite{1N4148_1,1N4148_2}.

The first approach, TBI (Section \ref{sec:spline-dev}), is a local parametric
regression technique that uses cubic splines to construct a piecewise polynomial
approximation of available electrical data \cite{deBoor,Gupta_17_INPROC}.  TBI
is used extensively in many modeling and simulation contexts, including compact
semiconductor device modeling, where it offers simplicity, computational
efficiency, and the ability to generate differentiable approximations. The
drawbacks of table-based models include significant memory requirements and 
datasets restricted to rectangular grids. We refer to \cite{Gupta_17_INPROC}
and \cite{Gupta_18_THESIS} for relevant recent work. 

The second approach (Section \ref{sec:GMLS-dev}), uses GMLS approximants
\cite{Wendland_04_BOOK} to build compact device models; this method, unlike
TBI, can be applied to scattered data as well.\footnote{Such datasets result
from scattered electrical measurements of devices with more than two terminals,
which will be considered in forthcoming work.} GMLS is an example of
non-parametric regression, which uses local kernels to build estimates from
scattered data. Scientific computing applications of GMLS  range from the
design of meshfree discretizations for PDEs \cite{Chen_17_JEM} to data
transfers for coupled multiphysics simulations
\cite{Slattery_16_JCP,Bungartz_16_CF}. We believe that we are the first to
apply GMLS to compact device modeling; in this paper, we not only develop
GMLS-based device models but also demonstrate them in circuit simulations. 

Finally, in Section~\ref{sec:DNN-dev}, we develop DNN \cite{Goodfellow_16_BOOK}
device models. DNNs are compositions of non-linear activation functions and
affine transformations, and represent global non-linear parametric
regression.  The success of DNNs in various classification tasks is well
documented \cite{LeCun_15_Nature}. Their application to scientific computing is
more recent \cite{Bar-Sinai_19_PNAS,Raissi_17_ArXiv} but is generating
significant interest. It should be noted that DNN applications to circuit
simulations
\cite{Zaabab_94_IEEE_MTT-S,Zaabab_95_IEEE_TMTT,Meijer_96_THESIS,Litovski_97_SPT,Andrejevic_03_JAC,Chen_06_INPROC,Gorissen_09_NCA}
predate these efforts, but have stayed fairly dormant over the years. It is
likely though that this research direction will intensify and attract more
attention, as evidenced by recent work \cite{Chen_17_INPROC}.  At the same
time, compact DNN models of devices are few and far between in the literature.
Early examples include \cite{Meijer_96_THESIS}, \cite{Zaabab_97_IEEE_TMTT},
and \cite{Hammouda_08_AJAS}, where DNNs were used to model various metal oxide
and field effect transistors. More recent work includes a multi-layer
perceptron (MLP) model of a transistor device \cite{Lei_18_INPROC}, and a
compact model for a thin TFET device using a hybrid MLP architecture with two
different activation functions \cite{Li_16_IEEE_JESSCDC}. 

Our work provides further insights into the development of DNN compact models,
informed by performing circuit simulations using such data-driven devices.   In
particular, our results strongly suggest that a ``reasonable'' Mean Square Error
(MSE) fit of $I\!\!-\!\!V$ characteristic curves alone may not be enough to
ensure convergence of a data-driven device in circuit simulations and/or
physically correct simulation results; for this, the compact model should also
possess actual device physics properties, such as passivity, monotonicity, zero
current at zero voltage,~\eetc{}. A key contribution of this paper is the
development of a DNN training strategy, based on transformed sets of electrical
measurements, that consistently produces physically correct compact diode models
across a range of DNN architectures; these models perform robustly in circuit
simulations, and produce results that are in excellent agreement with laboratory
measurements.

The rest of the paper is organized as follows. Section~\ref{sec:prelim} provides
some background information about the technical approach and the software tools
used in this work. Section~\ref{sec:dev} describes the core techniques
underlying the three regression methods above. Section~\ref{sec:results}
presents simulation results; to assess the performance of data-driven
compact device models, we first compare their  $I\!\!-\!\!V$ characteristics
with laboratory measurements using three different data views that expose
different aspects of device operation. Then, we use these compact models to build
a full-wave bridge rectifier circuit, simulate the circuit, and compare
simulation results against laboratory measurements. In
Section~\ref{sec:conclude}, we discuss our conclusions and outline directions
for future research. 

\section{Preliminaries}
\label{sec:prelim}

\subsection{Workflow}
\label{sec:workflow}

\begin{figure}[htbp!]
\centering
\includegraphics[width=0.9\textwidth]{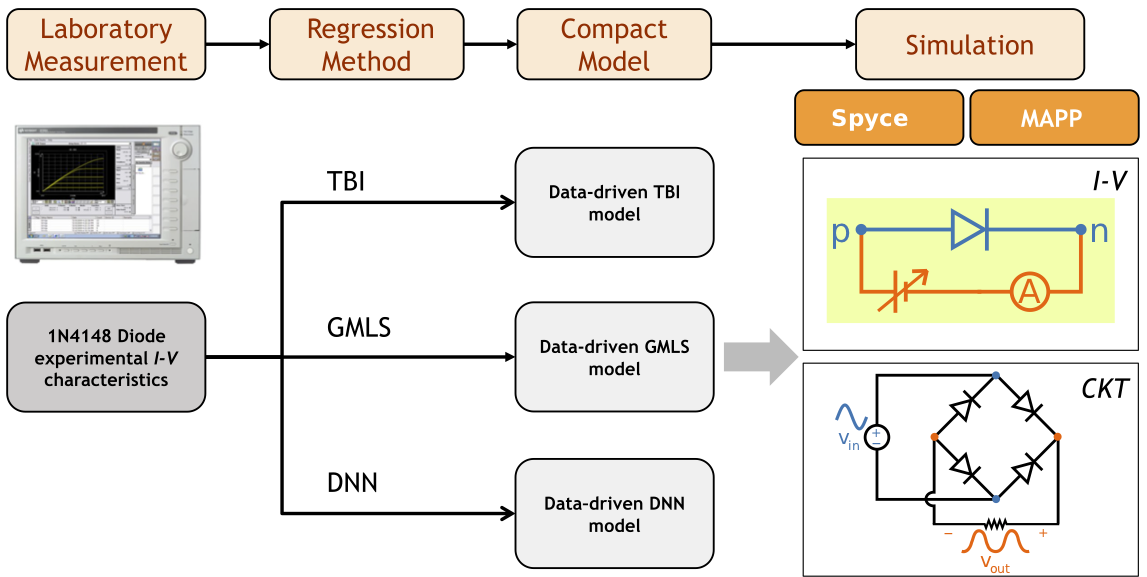}
\vspace{-2ex}
\caption{\small Our workflow to develop and test data-driven 
compact models, illustrated for the 1N4148 diode.} 
\label{fig:workflow}  
\vspace{-2ex}
\end{figure}

The main focus of this paper is the development and testing of data-driven
compact models based on three different regression approaches, exemplified using
a 1N4148 high-speed switching diode. Figure \ref{fig:workflow} shows the steps
involved in our compact model development and testing workflow, as applied to
this diode. 

As shown in the figure, we first obtain $I\!\!-\!\!V$ electrical measurements 
in the lab. Then, we apply our three different regression methods to this data, 
thereby generating three different sets of data-driven compact models for the 
device. We then simulate these data-driven compact models to obtain their 
$I\!\!-\!\!V$ characteristics, and we also simulate circuits where such models 
are deployed. Circuit simulations are then compared against laboratory 
measurements. 

\textit{Software: }We have implemented the entire workflow of
Figure~\ref{fig:workflow} both in MATLAB$\textregistered{}$ and in Python. For
TBI, we use two tools~(1)~STEAM \cite{Gupta_17_INPROC, Gupta_18_THESIS}, an
open-source MATLAB$\textregistered{}$ tool developed at UC Berkeley, and~(2)~a
Python implementation of cubic splines \PBB{developed for this work}. For GMLS, we use the open-source Compadre toolkit
\cite{Kuberry_19_MISC}, available as a Python package. For DNNs, we use
TensorFlow \cite{Tensorflow_15_MISC}, an open-source tool available as a Python
library. And for compact model and circuit simulations, we use:~(1)~the Berkeley
Model and Algorithm Prototyping Platform (MAPP) \cite{Wang_16_IEEE_MTT-S}, an
open-source circuit simulator written in MATLAB$\textregistered{}$, and (2)
Spyce, a Python-based \PBB{research} circuit simulator \PBB{developed at Sandia National Laboratories}.

\subsection{A compact p-n junction diode model}
\label{sec:PN-model}

A p-n junction diode has two terminals labeled $p$ and $n$, with $n$ serving as
the ``reference'' terminal. The voltage difference between $p$ and $n$ is
denoted $v_{pn}$, and the current flowing into the diode at the non-reference
terminal $p$ is denoted $i_{pn}$. The current flowing into the reference
terminal is $i_{np} = -i_{pn}$. A compact diode model is a mapping
$f_{PN}:\mathbb{R}\rightarrow\mathbb{R}$ that gives the diode current as a
function of the applied voltage, \ie{}, $i_{pn} = f_{PN}(v_{pn})$. Constructing
a compact diode model thus boils down to specifying the function $f_{PN}$. In
addition, the derivative of this function with respect to $v_{pn}$ is required 
by the non-linear solver in the circuit simulator. 

\begin{remark}
\label{rem:derivatives}
In traditional compact models, $f_{PN}$ is usually given by an analytic
expression and a set of $k$ parameters $\maxvec{p}=(p_1,\ldots,p_k)$ that
represent physical constants and/or variables that can be used to  calibrate
$f_{PN}$ to data. As a result, the derivative of  $f_{PN}$ can be obtained by
automatic differentiation (AD), and does not have to be provided as part of
the model \cite{Griewank2008EDP}. A typical example is the Shockley diode 
equation \cite{shockley1949theory},

\begin{equation}
    \label{eq:ideal}
    i_{pn} = f_{PN}(v_{pn}, \underbrace{(I_S,V_T,q)}_{\maxvec{p}}) = I_S \left( e^{\frac{v_{pn}}{q V_T}}-1\right).
\end{equation}

Here, $I_S$ is the reverse bias saturation current, $V_{T}$ is the thermal
voltage, and $q$ is the ``quality factor'', a non-physical parameter used to
account for imperfect p-n junctions in real diodes. AD may also be applied to 
some parametric regression models for $f_{PN}$, such as TBI. However, AD is not 
applicable to non-parametric regression such as GMLS, in which case the compact 
model must also provide its derivative. As a result, the accuracy of the 
regression fit for $f_{PN}$ alone is not enough to ensure the quality of such 
models; testing them in actual circuits should be an integral part of the 
development and validation process.
\end{remark}

In contrast to a traditional compact model such as \eqref{eq:ideal}, data-driven
models estimate $f_{PN}$ and its derivative by applying a regression technique
$\mathcal{R}$ to a dataset $\mathcal{D}_{PN} =
\{{v}^k_{pn},i^k_{pn}\}_{k=1}^m$. The dataset contains measurements of the
device's $I\!\!-\!\!V$ characteristic curve and corresponds to an $m\times 2$
matrix of real numbers.

\subsection{$I\!\!-\!\!V$ measurements of a 1N4148 diode}
\label{sec:data}

\begin{figure}[htbp!]
\centering
\includegraphics[width=0.85\textwidth]{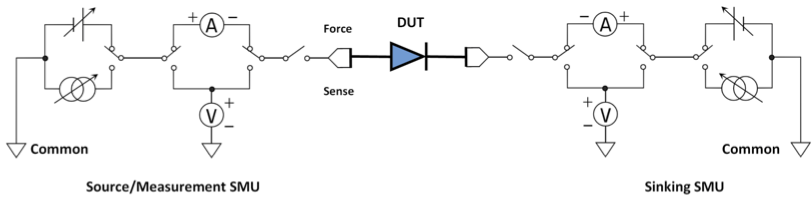}
\vspace{-2ex}
\caption{\small Schematic of our laboratory measurement setup for the 1N4148 diode.}
\label{fig:ckt-lab-measurement}  
\end{figure}

The $I\!\!-\!\!V$ characteristic of a device is a basic set of electrical
measurements and a fundamental way to understand the performance of various
materials and devices under test (DUT). $I\!\!-\!\!V$ measurements obtain the
current \vs{} voltage characteristic (denoted $\mathcal{D}_{PN}$ above) of a
device by applying a series of voltage stimuli to the device and measuring the
resulting current responses. For this work, we used a Keysight B1505A Parametric
Analyzer on a 1N4148 diode specimen to obtain $\mathcal{D}_{PN}$. This
parametric analyzer uses Source Measurement Units (SMUs) that combine a current
source, a voltage source, an ammeter, and a voltmeter into a single unit; see 
Figure \ref{fig:ckt-lab-measurement} for a schematic. We chose the B1505A's 
HPSMU (High Power Source Measurement Unit) for its ability to supply ample 
current over an extended voltage range while maintaining adequate 
measurement resolution.
 
For this work we sampled the  $I\!\!-\!\!V$ curve at $m=9682$ points, resulting
in a dataset $\mathcal{D}_{PN}$ given by a $9682\times 2$ matrix of real
numbers. The voltage stimuli ranged from $v_{pn}^{\min} = -125\textnormal{V}$
to $v_{pn}^{\max}=0.8\textnormal{V}$, in increments $\Delta v_{pn}$ ranging 
between $10\textnormal{mV}$ and $20\textnormal{mV}$. The non-uniformity of the 
voltage increments is due to rounding errors and the fact that the measurements
have inherent noise and stability issues, and are at the limit of forced voltage
step resolution.

\section{Core techniques for developing data-driven compact models}
\label{sec:dev}

In this section, we apply three different regression techniques $\mathcal{R}$ to
the dataset $\mathcal{D}_{PN}$ above, to develop three different sets of 
data-driven compact device models for the 1N4148 diode. 
 
\subsection{\PBB{Table-Based Interpolation} (TBI) devices}
\label{sec:spline-dev}

A TBI diode compact model is a function $i_{pn} = f_{PN}(v_{pn})$ that smoothly
interpolates the data points defined by the rows of $\mathcal{D}_{PN}$. There
are many ways to construct such a function~--~including cubic splines, Chebyshev
polynomials, and Barycentric Lagrange interpolation
\cite{deBoor,trefethen2013approximation}. In this paper, we use cubic splines;
they are simple to describe and construct, they produce robust, $C^2$-regular
compact models that converge well in circuit simulations, and they offer
inexpensive compact model evaluation as well as derivative computation as they
only require cheap univariate cubic polynomial evaluation (with pre-computed
coefficients that can be stored and looked up very efficiently). Examples of
such cubic spline driven TBI compact models can be found in
\cite{Gupta_17_INPROC} and \cite{Gupta_18_THESIS}.

\begin{remark}\label{rem:multi}
A table-based model for devices with more than two terminals involves
multivariate cubic spline interpolation. The main drawback of such models is
that they require electrical data sampled over a rectangular grid of voltages
and cannot be easily extended to scattered electrical data.  This is not an
issue for the univariate splines considered in this work.
\end{remark}

Below, we briefly review the construction of the univariate cubic splines used 
in this work. To improve convergence of the resulting compact models in circuit 
simulations, our development differs in important ways from standard splines 
found in the literature such as natural splines. 

To declutter notation, just for this subsection we switch to labeling the data
points in $\mathcal{D}_{PN}$ as $(x_{k}, y_{k})$, $k=1,\ldots,m$. Thus, we have
that $x_1 = v_{pn}^{\min}$, $x_m=v_{pn}^{\max}$, and $x_{k}=v_{pn}^k$ for
$1<k<m$. Likewise, $y_{k} = i^k_{pn}$. Without loss of generality, we assume 
that  $x_{1} < x_{2} < \ldots{} < x_{m}$.

The voltage measurements $x_{k}$ induce a partition of the $x$-axis into $m+1$
intervals  $(-\infty, x_{1})$, $[x_{1}, x_{2})$, $[x_{2}, x_{3})$, $\ldots$,
$[x_{m-1}, x_{m})$, $[x_{m}, +\infty)$. We refer to the first and last interval
as ``boundary'' intervals and the rest as ``interior'' intervals. A cubic spline
defined with respect to this partition is a piecewise cubic polynomial $f(x)$
which interpolates the  data $(x_k,y_k)$, $k=1,\ldots,m$, and has continuous
first and second derivatives, \ie{}, it is of class $C^2(\mathbb{R})$. We denote
the restriction of $f(x)$ to the $i^{th}$ interval above as $C_{i}(x) = a_{i}
x^{3} + b_{i} x^{2} + c_{i} x + d_{i}$, for $1 \leq i \leq m+1$. Succinctly, 

\begin{equation*}
    f(x) = \begin{cases}
           C_{1}(x) \textnormal{, if } x < x_{1},\\
           C_{i+1}(x) \textnormal{, if } x \in [x_{i}, x_{i+1}) \textnormal{ for each } 1 \leq i \leq m-1, \textnormal{ and}\\
           C_{m+1}(x) \textnormal{, if } x \geq x_{m}.
           \end{cases}
\end{equation*}

To determine the $4(m+1)$ polynomial coefficients
$\{a_{i},b_{i},c_{i},d_{i}\}_{i=1}^{m+1}$ defining the cubic spline segment on
each interval, we enforce the following $4(m+1)$ constraints:

\begin{itemize}
\setlength{\itemsep}{0.1ex}
    \item{$C_{i}(x_{i}) = y_{i}$ and $C_{i+1}(x_{i}) = y_{i}$; $1 \leq i \leq m$: interpolation ($2m$ constraints).}
    \item{$C_{i}^{'}(x_{i}) = C_{i+1}^{'}(x_{i})$; $1 \leq i \leq m$: continuity of first derivatives ($m$ constraints).}
    \item{$C_{i}^{''}(x_{i}) = C_{i+1}^{''}(x_{i})$; $1 \leq i \leq m$: continuity of second derivatives ($m$ constraints).}
    \item{$a_{1} = b_{1} = 0$ and $a_{m+1} = b_{m+1} = 0$: linearity at boundary intervals ($4$ constraints).}
\end{itemize}

These conditions are sufficient to determine a unique, globally $C^2$ piecewise
cubic interpolant $f(x)$ of the data in $\mathcal{D}_{PN}$. In practice, given
the electrical measurements $\mathcal{D}_{PN}$, the constraints above are used
to pre-compute and store the coefficient set
$\{a_{i},b_{i},c_{i},d_{i}\}_{i=1}^{m+1}$ in memory.

To evaluate the compact model, \ie{}, to compute $f_{PN}$ and $f^\prime_{PN}$ at
a query point $v^*_{pn}$, one first locates the interval containing
$v^*_{pn}$ and retrieves the four coefficients corresponding to the restriction
of $f_{PN}$ to that interval. Then the corresponding cubic polynomial and its
derivative are calculated and returned.

\subsection{GMLS devices}
\label{sec:GMLS-dev}

GMLS is a non-parametric regression approach for approximating linear
functionals from scattered data \cite{Wendland_04_BOOK}. Here, we use GMLS to
estimate $f_{PN}$ and $f^\prime_{PN}$ from the data $\mathcal{D}_{PN}$. Below, 
we describe the basics necessary for this task and refer to 
\cite{Wendland_04_BOOK} for further details.

Consider a $C^1$ function ${f}: \boldmath{R}\rightarrow \boldmath{R}$ with
domain $\mathcal{D}\subseteq \mathbb{R}$, a point cloud $X$ of size $m$, 
$X=\{{x}_1,\ldots,{x}_m\}\subset \mathcal{D}$ with a fill distance $h_X$, and a collection of
point samples $\mathbf{f}=\{{f}({x}_i)\}_{i=1}^{m}$ on the cloud. Let
$P_k(\mathbb{R})$ denote the set of all real polynomials of degree less than or
equal to $k>0$ with dimension $\dim P_k(\mathbb{R})=k+1$ and basis $\bm{\phi}=\{
\phi_q\}_{q=1}^{k+1}$. Finally, $\rho_\epsilon (r)$ will denote a radially
symmetric kernel function that is at least $C^1(\mathbb{R})$ and whose support
is contained in $(-\epsilon, \epsilon)$ for some real $\epsilon >0$.  

Given a point ${x}^\star\in\mathcal{D}$, GMLS computes approximations
$\widetilde{{f}}({x}^\star)\approx {f}({x}^\star)$ and $\widetilde{{d_x}
{f}}({x}^\star)\approx {d_x} {f}({x}^\star)$ that are exact for all ${f}\in
P_k(\mathbb{R})$ (polynomial reproduction) in two steps. To describe these
steps, let  $W({x}^\star) \in \mathbb{R}^{m\times m}$ be a diagonal matrix with
element $W_{ii}({x}^\star)=\rho_\epsilon(|{x}^\star - {x}_i |)$. Let $B \in
\mathbb{R}^{m\times (k+1)}$, the  matrix with element $ B_{ij} = \phi_j({x}_i)$;
$i=1,\ldots, m$;  $j=1,\ldots,k+1$. And let $\|\cdot\|_{W}$ be the Euclidean
$\ell^2$ norm on $\mathbb{R}^m$ weighted by $W({x}^\star)$. The two GMLS
steps are: \smallskip

\emph{Step 1. Computing the GMLS coefficient vector.} Solve the weighted least-squares problem
\begin{equation}\label{eq:gmls-b}
\mathbf{c}(\mathbf{f})
= \argmin_{\bm{c}\in\mathbb{R}^{k+1}}\frac12
\left \| B\mathbf{c} -  \mathbf{f}  \right \|^2_{W}\,.
\end{equation}
%

\emph{Step 2. Computing the GMLS approximants.}  Set\footnote{The GMLS
derivative approximation in \eqref{eq:GMLS-approx} \emph{appears} to violate the
product rule and for this reason it was often referred to as the ``diffuse
derivative approximation'' in the literature; see \cite{Mirzaei_12_IMAJNA}.
This confusion stems from misconstruing how the GMLS approximation works and
assuming (erroneously) that $\widetilde{{d_x} {f}}({x}^\star)$ is defined by
differentiating $\widetilde{{f}}({x}^\star)$. In fact, the GMLS derivative
approximation is derived independently of the GMLS field approximation and does
not involve differentiation of the latter; see \cite{Wendland_04_BOOK}.} 

\begin{equation}\label{eq:GMLS-approx}
\widetilde{{f}}({x}^\star) := \bm{c}(\mathbf{f})\cdot \bm{\phi}({x}^\star)
\quad\mbox{and}\quad 
\widetilde{{d_x} {f}}({x}^\star):=\bm{c}(\mathbf{f})\cdot {d_x} \bm{\phi}({x}^\star)\,.
\end{equation}

To evaluate the GMLS compact diode model at a given query point $v^*_{pn}$,
we proceed as follows. If $v^*_{pn}\in[v_{pn}^{\min},v_{pn}^{\max}]$, we
associate the first and the second columns of $\mathcal{D}_{PN}$ with a point
cloud $X\subset\mathbb{R}$ and a sample set $\mathbf{f}\in\mathbb{R}^m$,
respectively, solve \eqref{eq:gmls-b} and define $f_{PN}(v^*_{pn})$ and
${d_x}f_{PN}(v^*_{pn})$ according to \eqref{eq:GMLS-approx}. If
$v^*_{pn}\notin[v_{pn}^{\min},v_{pn}^{\max}]$, we evaluate
$f_{PN}(v^*_{pn})$ and ${d_x}f_{PN}(v^*_{pn})$ as follows. Let
$v^{\text{clo}}_{pn}$ be the point from  $\mathcal{D}_{PN}$ that is  closest to
$v^*_{pn}$. Note that $v^{\text{clo}}_{pn}$ is either $v_{pn}^{\min}$ or
$v_{pn}^{\max}$. We then set $$ f_{PN}(v^*_{pn}):=f_{PN}(v^{\text{clo}}_{pn}) +
(v^*_{pn} - v^{\text{clo}}_{pn}){d_x}f_{PN}(v^{\text{clo}}_{pn})
\quad\mbox{and}\quad {d_x}f_{PN}(v^{*}_{pn}):={d_x}f_{PN}(v^{\text{clo}}_{pn}),
$$ respectively,  where $f_{PN}(v^{\text{clo}}_{pn})$ and
${d_x}f_{PN}(v^{\text{clo}}_{pn})$ are the GMLS estimates at
$v^{\text{clo}}_{pn}$. 

In this work, we use the Compadre toolkit \cite{Kuberry_19_MISC} for performant
implementation of GMLS, with polynomial orders $k=1,2,3$, and kernel $\rho(r) =
\left(1-{r}/{\varepsilon}\right)^p_+$, with $p = 4$ and $\varepsilon=h_X$.
Compadre uses an adaptive procedure to adjust $\varepsilon$ until
$\mbox{supp}(\rho_\epsilon(|{x}^\star - {x}_i |)\cap X$ is guaranteed to contain
enough points to ensure the desired degree of polynomial reproduction. For real
polynomials and quasi-uniform point clouds, the number of points selected by
this procedure does not exceed $2(k+1)$.

Compadre solves the weighted least-squares problem \eqref{eq:gmls-b} using QR
factorization, which formally requires $2(k+1)^2 (m - (k+1)/3)$ flops
\cite[p.240]{Golub_96_BOOK}. However, the actual number of non-zero rows in $B$
equals the number of points selected by Compadre's adaptive procedure and is
bounded by $2(k+1)$, $k=1,2,3$, \ie{}, it is orders of magnitude less than the
size $m$ of the dataset $\mathcal{D}_{PN}$. As a result, the actual cost per
model evaluation is $O((k+1)^3)$ rather than $O((k+1)^2 m)$. In practice, we did
not observe noticeable differences in the performance of GMLS compact models
as we increased the polynomial degree from 1 to 3. 

\subsection{DNN devices}
\label{sec:DNN-dev}

In this work, we use a simplified DNN definition from
\cite{Opschoor_19_TECHREPORT}. Let $d$ and $D$ be two natural numbers defining
the input dimension and the depth of the network, respectively.  Consider a set
of natural numbers $\{n_0,\ldots, n_D\}$ such that $n_0=d$, and a set of
matrix-vector tuples $\{A_i,\mathbf{b}_i\}$, $i=1,\ldots,D$ such that $A_i \in
\mathbb{R}^{n_i\times n_{i-1}}$ and $\mathbf{b}_i\in \mathbb{R}^{n_i}$. The
elements of $A_i$ and $\mathbf{b}_i$ are usually called the weights and the
biases of the DNN respectively. Finally, let $\sigma:\mathbb{R}\rightarrow
\mathbb{R}$ be a non-linear ``activation'' function. The action of the resulting
DNN on an input vector $\mathbf{x}\in\mathbb{R}^d$ is defined as
\begin{equation}\label{eq:DNN}
\left\{
\begin{array}{rcl}
\mathbf{y}_0 &:=& \mathbf{x} \\
\mathbf{y}_k &:= & \sigma(A_k\mathbf{y}_{k-1} + \mathbf{b}_k)\quad\mbox{for $k=1,\ldots, D-1$} \\
\mathbf{y}_D &:=& A_D\mathbf{y}_{D-1} + \mathbf{b}_D
\end{array}
\right. \,,
\end{equation} 
where $\sigma$ is applied component-wise. We denote the transformation of the
input vector by the DNN as $\mathcal{N}(\mathbf{x})$, that is, $\mathbf{y}_D =
\mathcal{N}(\mathbf{x})$.  The mapping $\mathcal{N}:\mathbb{R}^d\rightarrow
\mathbb{R}^{n_D}$ defines a global non-linear function parameterized by the
unknown weights and biases $\{A_i,\mathbf{b}_i\}$.  To determine these
parameters one ``trains'' the network by solving a constrained optimization
problem,
\begin{equation}\label{eq:opt}
\{A_i,\mathbf{b}_i\} = \argmin \mathcal{L}(\mathcal{N}(\mathbf{X}_T),\mathbf{Y}_T)
\quad\mbox{subject to}\quad \mathcal{C}\left(\{A_i,\mathbf{b}_i\}\right)
\end{equation}
where $\mathcal{L}$ is a ``loss'' function measuring the mismatch between
the network's output and the training output, $\mathcal{C}$ is a non-linear
constraint operator, $\mathbf{X}_{T} = \{\mathbf{x}_{T,1},\ldots,
\mathbf{x}_{T,m}\}$ is a set of training inputs, and $\mathbf{Y}_{T} =
\{\mathbf{y}_{T,1},\ldots, \mathbf{y}_{T,m}\}$ are the corresponding outputs. We
refer to the pair $\{\mathbf{X}_{T}, \mathbf{Y}_{T}\}$ as the training set and
denote it by $\mathcal{T}$. In this work, we train the neural network model
using Adam, a variant of stochastic gradient descent, with a maximum of $E$ epochs.

For classification tasks, the non-linearity of $\mathcal{N}$ is used to
transform input datasets representing different classes into linearly separable
sets.  In contrast, here we shall use \eqref{eq:DNN} as a \textit{regression}
tool to build a compact diode model from $I\!-\!V$ measurements.  To that end,
we set the input dimension $n_0=1$, the output dimension $n_{D}=1$ and define
$\mathcal{L}$ as the mean square (MSE) error, a common choice for regression
tasks.  Furthermore, since circuit  simulations require differentiable compact
models, we only consider smooth activation functions $\sigma$ and forego the
widely used piecewise linear ReLU activation. As a constraint operator
$\mathcal{C}$, we use non-negativity of the weights to ensure monotonicity of
the function $f_{PN}$, as required by the physics of p-n junction diodes. 

We train the network using three different options for the training set
$\mathcal{T}$. The first one is $\mathcal{T}:=\mathcal{D}_{PN}$, \ie{}, the
original set of $I\!-\!V$ measurements with bounding box
$[v^{\min}_{pn},v^{\max}_{pn}]\times [i^{\min}_{pn},i^{\max}_{pn}]$. The second
option is the \emph{transformed} dataset  $\mathcal{D}^{VI}_{PN}:= \{
T_V(v^k_{pn}),T_{I}(i^k_{pn})\}$, where 
\begin{equation}\label{eq:transform-V}
T_V (v_{pn}) =
\left\{
\begin{array}{rl}
v_{pn}/V^{+} & \mbox{if $v_{pn}\ge 0$},\\   
v_{pn}/V^{-} & \mbox{if $v_{pn}<0$}       
\end{array}
\right. \,\quad\mbox{and}\quad
\end{equation}
\begin{equation}\label{eq:transform-I}
T_I(i_{pn}) =
\left\{
\begin{array}{rl}
-(\log_{10}(-i_{pn}) - B^{-})/A^{-} &  \mbox{if $i_{pn} \in (-\infty,-P^{-}_{\min}] $} \\
-1 + S(i_{pn} + P^{-}_{\min}) & \mbox{if  $i_{pn}\in(-P^{-}_{\min}, P^{+}_{\min})$} \\
(\log_{10}( i_{pn}) - B^{+})/A^{+} &  \mbox{if $i_{pn} \in [P^{+}_{\min},\infty) $} 
\end{array}
\right. \,;
\end{equation}
with $P^{\pm}_{\min} = 10^{p^{\pm}_{\min}}$, $A^{\pm} = (p^{\pm}_{\max} -
p^{\pm}_{\min})/7$, $B^{\pm} =p^{\pm}_{\min}-A^{\pm}$, and $S=
2/(P^{+}_{\min}+P^{-}_{\min})$. 
The last option is the \emph{partially transformed} set
$\mathcal{D}^{I}_{PN}:=\{v^k_{pn},T_{I}(i^k_{pn})\}$, \ie{}, only $i_{pn}$ is
transformed, not $v_{pn}$.  
For the numerical examples, we set  $V^{+} = 0.1$, $V^{-} = 15.625$,
$p^{\pm}_{\min} = -10$, $p^{+}_{\max} = -1$ and $p^{-}_{\max}=-5$. 
Given a training set $\mathcal{T}^{*}_{PN}$ the training process yields an
instance of \eqref{eq:DNN} which we label as $\mathcal{N}^{*}_{PN}$. The
data-driven compact device model is then defined as 
$$
f_{PN}(v_{pn}) = 
\left\{
\begin{array}{rl}
\mathcal{N}_{PN}(v_{pn}) 
& \mbox{if $\mathcal{T} = \mathcal{D}_{PN}$}
 \\ [1ex]
T^{-1}_{I}\circ\mathcal{N}^{VI}_{PN}\circ T_V(v_{pn}) 
& \mbox{if $\mathcal{T} = \mathcal{D}^{VI}_{PN}$} 
\\ [1ex]
T^{-1}_{I}\circ\mathcal{N}^{I}_{PN}(v_{pn}) 
& \mbox{if $\mathcal{T} = \mathcal{D}^{I}_{PN}$} 
\end{array}.
\right.
$$

\PBB{Training DNNs on appropriately transformed electrical measurement data is one of the key ideas in this paper. Our results in Section \ref{sec:results} show that this strategy consistently produces DNN compact models that provide accurate data fit and perform well in circuit simulations.} 


\PBB{The heuristics supporting the application of transformations is motivated
by}
the physics of p-n junction diodes: when $v_{pn} < 0$, $i_{pn}$ is a very small,
negative current (in our dataset, this current ranged from about $-0.5$
nano-Amperes to $-1.5$ micro-Amperes). 
\PBB{To mitigate the inability of the MSE loss function to differentiate between
such small values, we transform the current in the reverse bias regime roughly
as $i_{pn}\to -log_{10}(-i_{pn})$.} \PBB{In so doing}, a negative nano-Ampere
current \PBB{transforms} into $9$ and a negative micro-Ampere \PBB{becomes} a
$6$. To preserve the negativity of the transformed currents, we shift them by a
constant negative value; for example, if the shift is $-10$, a negative
nano-Ampere and a negative micro-Ampere transform into $-1$ and $-4$
respectively. Such large values are easily resolvable by the MSE loss function,
and hence the resulting DNNs \PBB{tend} to match the training data more
accurately. 

\PBB{Conversely}, when $v_{pn} > 0$, $i_{pn}$ is usually a positive current than
takes on a much wider range of values (in our dataset, this current ranged from
$0.5$ pico-Amperes to $35$ milli-Amperes). To resolve these values \PBB{with the
MSE loss function}, we again apply a log transformation, followed by a shift so
that the positivity of $i_{pn}$ is preserved. \PBB{For instance,  with }a shift
of $14$, a pico-Ampere \PBB{becomes} a 2 and a milli-Ampere \PBB{transforms
into} an 11. 

However, since we have used different transformations for $i_{pn} < 0$ and
$i_{pn} > 0$, we will have a discontinuity at $i_{pn} = 0$, which can cause
convergence problems in circuit simulations. To get around this, we choose a
small interval around $i_{pn} = 0$, apply the respective transformations only
outside this interval, and apply a continuity-preserving linear transformation
within this interval. 
\PBB{This approach ensures that an MSE loss function in conjunction with the
transformed set $\mathcal{D}^{VI}_{PN}$ resolves both positive and negative p-n
junction currents very well, leading to a much better fit.}

Configuring and training a DNN involves choosing several parameters, such as the
number of hidden layers $D-1$, the number of neurons $n$ per hidden layer, the
activation function, whether or not to use a kernel constraint, whether or not
to transform $v_{pn}$, and whether or not to transform $i_{pn}$. In this work,
we experimented with DNNs containing $1$ and $2$ hidden layers, with $5$, $10$,
$25$, $50$, or $100$ neurons per hidden layer, and $3$ different activation
functions (eLU, sigmoid, and tanh). Taken together with the $3$ binary choices
of whether to have a kernel constraint, transform $v_{pn}$, and/or transform
$i_{pn}$, this corresponds to a parameter space of size $2 \times 5 \times 3
\times 2 \times 2 \times 2 = 240$. We systematically explored this space,
generating all $240$ DNN compact models. Then we simulated them all using Spyce,
to determine their $I\!\!-\!\!V$ characteristics as well as their performance in
a bridge rectifier circuit. We do not need to discuss all the DNNs here; a
carefully chosen representative sample of ``interesting'' DNNs (summarized in
Table~\ref{tab:DNN}) suffice to highlight the key points we wish to make. 


%
\begin{table}[h!]
\renewcommand{\arraystretch}{0.5}
\scriptsize
\centering
\textsf{
\begin{tabular}{
p{0.2\textwidth}
>{\centering}p{0.1\textwidth}
>{\centering}p{0.05\textwidth}
>{\centering}p{0.05\textwidth}
>{\centering}p{0.05\textwidth}
>{\centering\arraybackslash}p{0.1\textwidth}
}
\hline\noalign{\smallskip}
Model & $\sigma$ & $D-1$ & $n_i$ & $\mathcal{T}$ & $\mathcal{C}$\\
\hline\noalign{\smallskip}
M1-50-E &  eLU &  1 &  50 & $\mathcal{D}_{PN}$ & Non-neg\\
\noalign{\smallskip}\hline\noalign{\smallskip}
M2-50-E &  eLU &  2 &  50 & $\mathcal{D}_{PN}$ & Non-neg\\
\noalign{\smallskip}\hline\noalign{\smallskip}
M1-50-S &  sigmoid &  1 &  50 & $\mathcal{D}_{PN}$ & Non-neg\\
\noalign{\smallskip}\hline\noalign{\smallskip}
M2-50-S &  sigmoid &  2 &  50 & $\mathcal{D}_{PN}$ & Non-neg\\
\noalign{\smallskip}\hline\noalign{\smallskip}
M1-50-T &  tanh &  1 &  50 & $\mathcal{D}_{PN}$ & Non-neg\\
\noalign{\smallskip}\hline\noalign{\smallskip}
M1-50-S-neg &  sigmoid &  1 &  50 & $\mathcal{D}_{PN}$ & None\\
\noalign{\smallskip}\hline\noalign{\smallskip}
M2-50-S-neg &  sigmoid &  2 &  50 & $\mathcal{D}_{PN}$ & None\\
\noalign{\smallskip}\hline\noalign{\smallskip}
M1-10-E-VI & eLU & 1 & 10 & $\mathcal{D}_{PN}^{VI}$ & Non-neg\\
\noalign{\smallskip}\hline\noalign{\smallskip}
M2-10-E-VI & eLU & 2 & 10 & $\mathcal{D}_{PN}^{VI}$ & Non-neg\\
\noalign{\smallskip}\hline\noalign{\smallskip}
M1-10-E-VI-neg & eLU & 1 & 10 & $\mathcal{D}_{PN}^{VI}$ & None\\
\noalign{\smallskip}\hline\noalign{\smallskip}
M2-10-E-VI-neg & eLU & 2 & 10 & $\mathcal{D}_{PN}^{VI}$ & None\\
\noalign{\smallskip}\hline\noalign{\smallskip}
M1-10-T-VI & tanh & 1 & 10 & $\mathcal{D}_{PN}^{VI}$ & Non-neg\\
\noalign{\smallskip}\hline\noalign{\smallskip}
M2-10-T-VI & tanh & 2 & 10 & $\mathcal{D}_{PN}^{VI}$ & Non-neg\\
\noalign{\smallskip}\hline\noalign{\smallskip}
M2-25-T-VI & tanh & 2 & 25 & $\mathcal{D}_{PN}^{VI}$ & Non-neg\\
\noalign{\smallskip}\hline\noalign{\smallskip}
M2-50-T-VI & tanh & 2 & 50 & $\mathcal{D}_{PN}^{VI}$ & Non-neg\\
\noalign{\smallskip}\hline\noalign{\smallskip}
M2-100-T-VI & tanh & 2 & 100 & $\mathcal{D}_{PN}^{VI}$ & Non-neg\\
\noalign{\smallskip}\hline\noalign{\smallskip}
\end{tabular}
}
\caption{\small Summary of DNN architectures and training configurations reported in this work.}
\label{tab:DNN}
\end{table}

\section{Results}
\label{sec:results}

%
%
%
%

\PBB{In this section, we carry out two types} of simulations for each data-driven
diode compact model: we~(1)~produce its $I\!\!-\!\!V$ characteristics,
and~(2)~simulate a bridge rectifier circuit (schematic shown at the bottom right
of Figure \ref{fig:workflow}) that uses $4$ instances of the model. 
\PBB{The input signal to the rectifier is a sine wave with frequency $10$Hz and phase shift approximately $\pi/1.63$. This simulation provides device model validation at a circuit level.} We compare the results of both simulations against laboratory data. 
%
\begin{figure}[htbp!]
\centering
\includegraphics[width=0.85\textwidth]{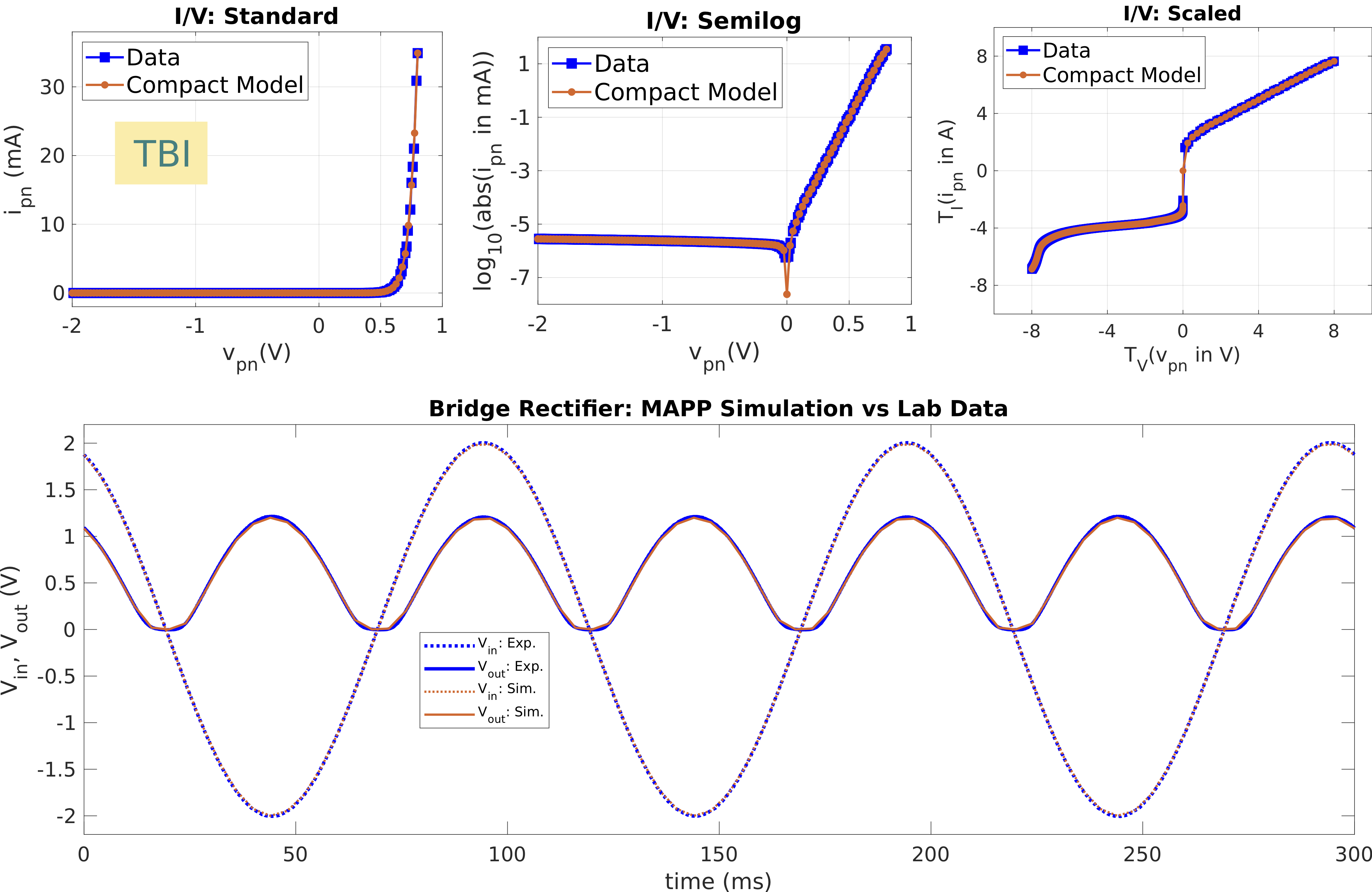}
\vspace{-1em}
\caption{\small Simulations comparing a cubic spline TBI diode compact model against laboratory data}
\label{fig:STEAM}
\end{figure}
%
%
\PBB{To visualize how well a data-driven diode compact model matches measured data,
we arrange results in plots containing $3$ top sub-plots and $1$ bottom sub-plot 
(see, for example, Figure~\ref{fig:STEAM}).}
The top sub-plots 
show the $I\!\!-\!\!V$ characteristics of the compact model, overlaid on top of
laboratory data, using three different ``data views" defined as follows:

\begin{itemize}
\setlength{\itemsep}{0.1ex}
\item{Standard (top-left sub-plot): $i_{pn}$ is plotted as a function of
      $v_{pn}$ for $-2V \le v_{pn} \le 0.8V$.}
\item{Semilog (top-middle sub-plot): $\log_{10}(abs(i_{pn}))$ is plotted as a
      function of $v_{pn}$ for $-2V \le v_{pn} \le 0.8V$.}
\item{Scaled (top-right sub-plot):  $T_I(i_{pn})$ is plotted as a function of
      $T_V(v_{pn})$, for $v^{\min}_{pn} \le v_{pn} \le v^{\max}_{pn}$, where 
      $T_V$ and $T_I$ are the transformations defined in \eqref{eq:transform-V} 
      and \eqref{eq:transform-I}, respectively.}
\end{itemize}

Together, the three data views \PBB{provide} a comprehensive picture of the model: the standard 
view \PBB{highlights} the \textit{forward bias} regime where the diode conducts
positive, exponentially-growing current $i_{pn}$ for positive $v_{pn}$. The 
semilog view highlights the ``zero crossing point'', where the diode 
transitions from \textit{reverse bias} (where it conducts very little current) 
to \textit{forward bias}. The scaled view exposes the entire region of
operation of the diode, from \textit{avalanche breakdown} (where $i_{pn}$ 
starts to become more and more negative at very low $v_{pn}$, risking
irreversible damage to the device) all the way to \textit{forward bias}.

The bottom sub-plot in each figure shows a bridge rectifier circuit simulation
using $4$ instances of the data-driven diode compact model (see schematic at the
bottom right of Figure \ref{fig:workflow}), overlaid on top of laboratory data.

\begin{figure}[htbp!]
\centering
\includegraphics[width=0.85\textwidth]{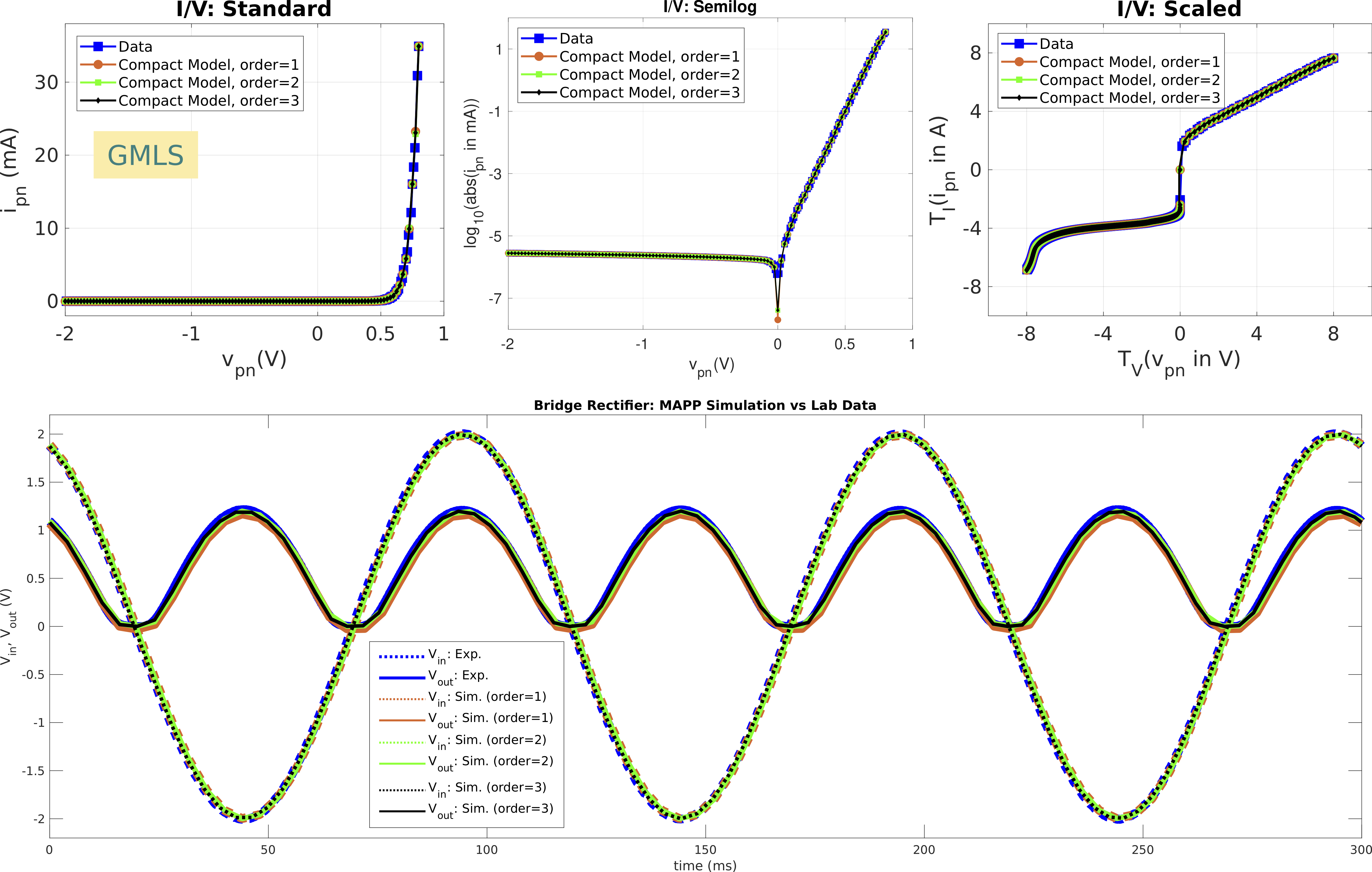}
\vspace{-1em}
\caption{\small Simulations comparing GMLS diode compact models against laboratory data}
\label{fig:GMLS}
\end{figure}

Figure \ref{fig:STEAM} shows simulation results for a cubic spline
TBI diode compact model, generated via STEAM and MAPP. Figure \ref{fig:GMLS}
shows the same for $3$ GMLS compact models (with polynomial orders
1, 2, and 3). In both cases, we see that the data-driven models are in excellent
agreement with laboratory data. We do not show results for TBI and 
GMLS models generated via Spyce, because they are virtually identical to the 
MAPP/STEAM results \PBB{shown here}.

We now turn to compact DNN models.  \PBB{Using the original dataset
$\mathcal{D}_{pn}$ as the training set failed to produce even a single accurate
DNN compact model. For all parameter choices described in
Section~\ref{sec:DNN-dev}, the learned DNN models were both inaccurate and
unphysical.} 

\begin{figure}[htbp!]
\centering
\includegraphics[width=0.85\textwidth]{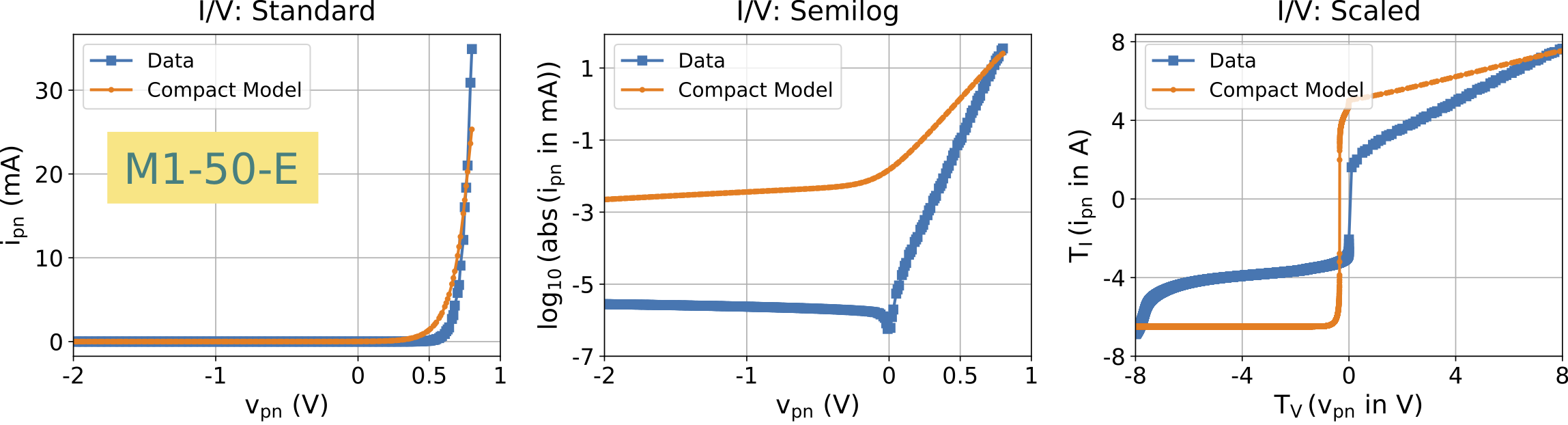}
\includegraphics[width=0.85\textwidth]{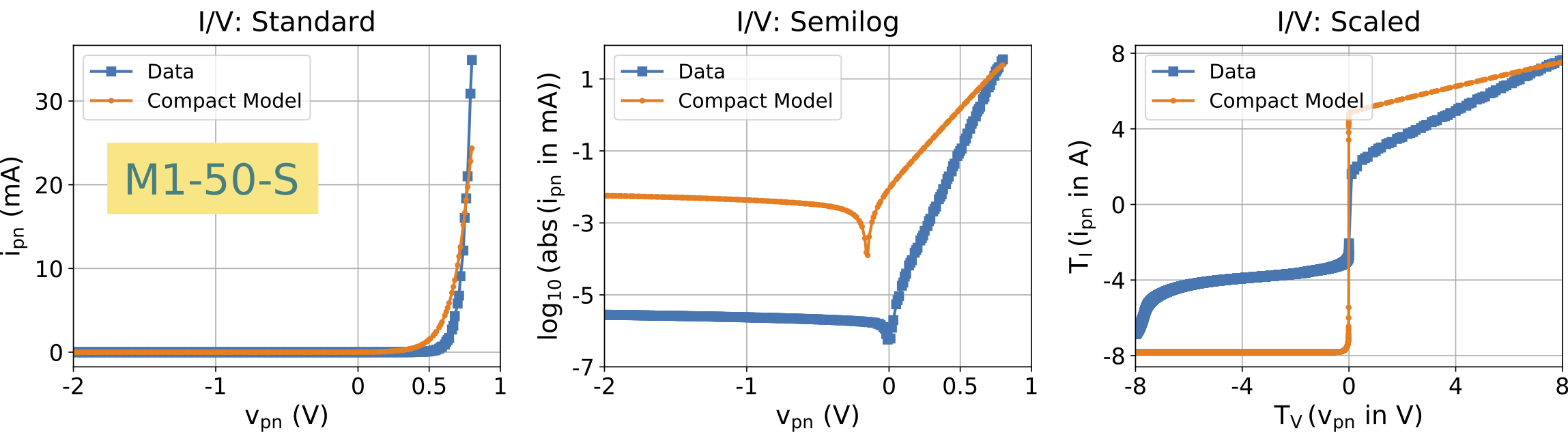}
\includegraphics[width=0.85\textwidth]{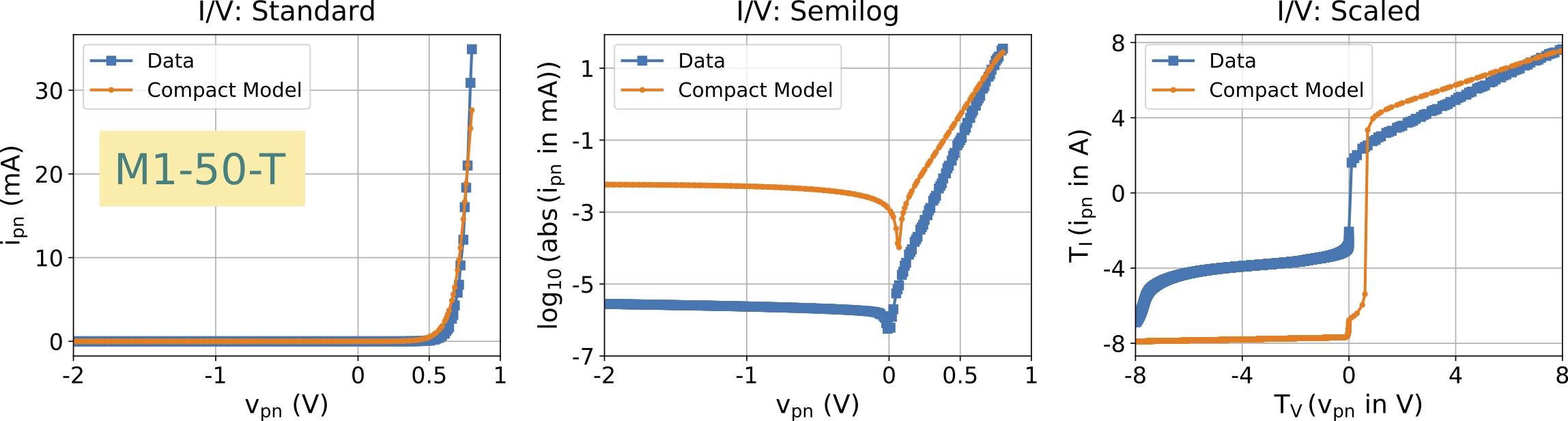}
\caption{\small Typical $I\!-\!V$ characteristics of DNN compact models learned 
without data transformations but with the non-negative kernel constraint 
enforced: M1-50-E (top), M1-50-S (middle) and M1-50-T (bottom). These results 
also highlight distinctions between eLU, sigmoid and tanh activation functions. 
}
\label{fig:notx-noty}
\end{figure}

\begin{figure}[htbp!]
\centering
\includegraphics[width=0.85\textwidth]{M1-50-E-noneg-notx-noty-PNG}
\includegraphics[width=0.85\textwidth]{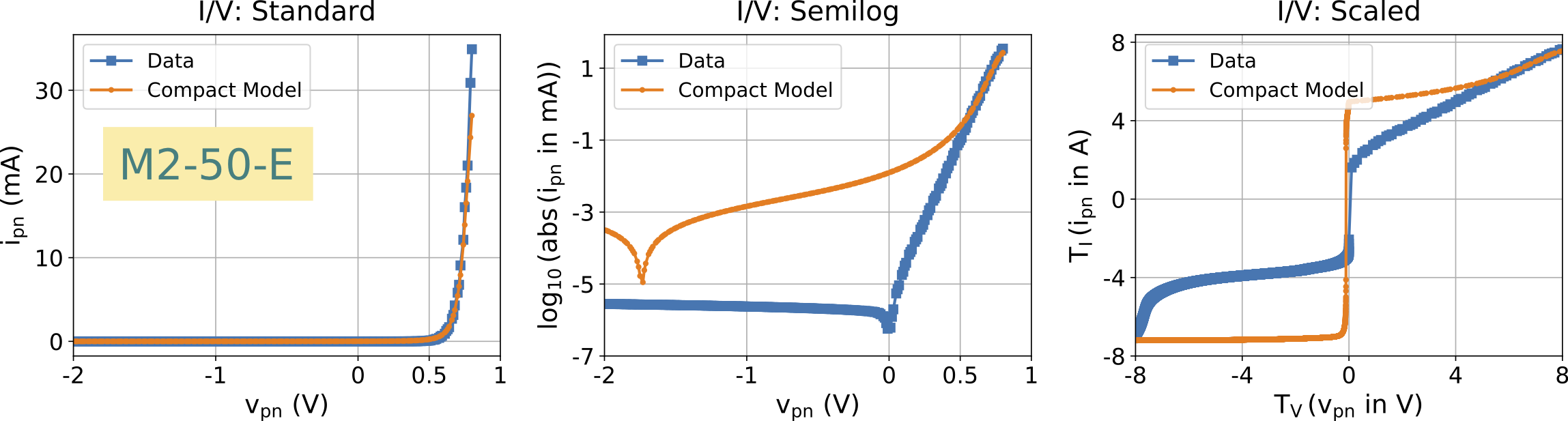}
\caption{\small Typical $I\!-\!V$ characteristics of DNN compact models learned without data transformations but with the non-negative kernel constraint enforced: M1-50-E (top), M2-50-E (bottom). These results show that increasing the depth can significantly change the regression fit without improving its quality. 
}
\label{fig:E-notx-noty-1vs2}
\end{figure}

\begin{figure}[htbp!]
\centering
\includegraphics[width=0.85\textwidth]{M1-50-S-noneg-notx-noty-PNG}
\includegraphics[width=0.85\textwidth]{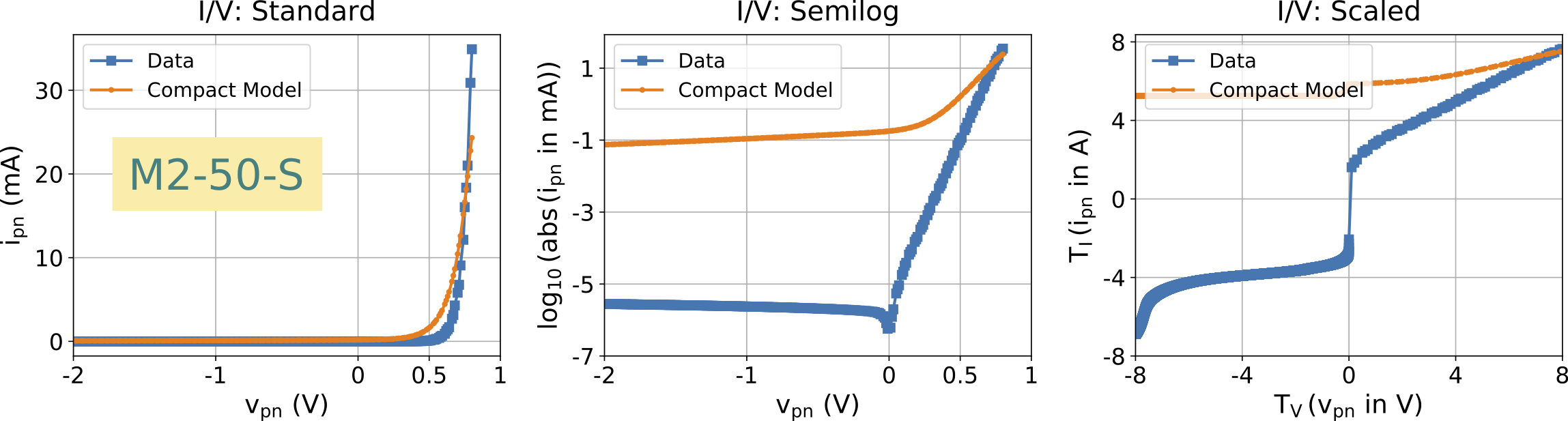}
\caption{\small Typical $I\!-\!V$ characteristics of DNN compact models learned without data transformations but with the non-negative kernel constraint enforced: M1-50-S (top), M2-50-S (bottom). These results show that increasing the depth can significantly change the regression fit without improving its quality.  
}
\label{fig:S-notx-noty-1vs2}
\end{figure}

\begin{figure}[t!]
\centering
\includegraphics[width=0.85\textwidth]{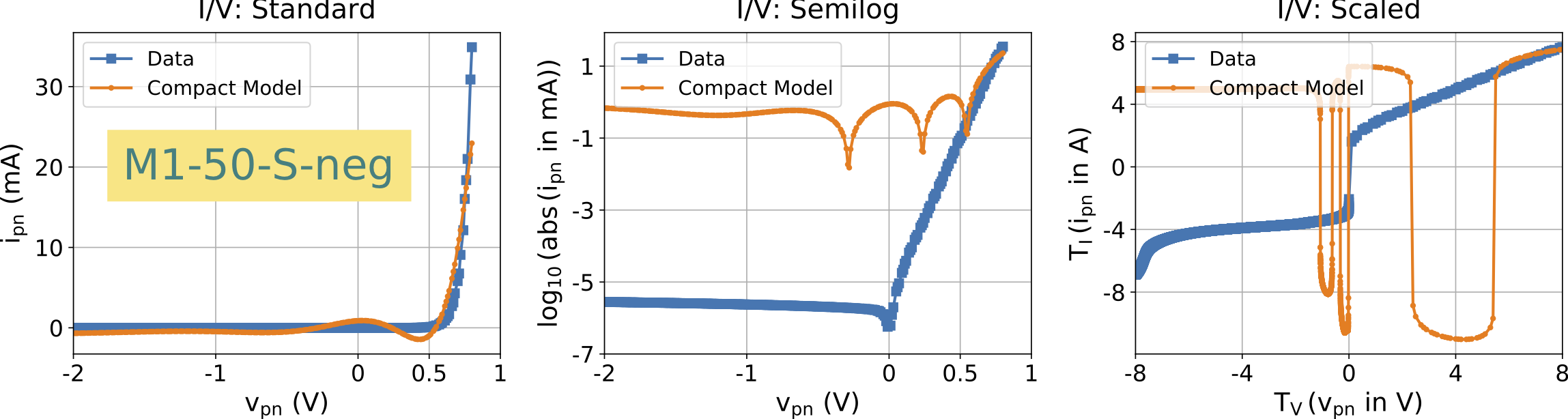}
\includegraphics[width=0.85\textwidth]{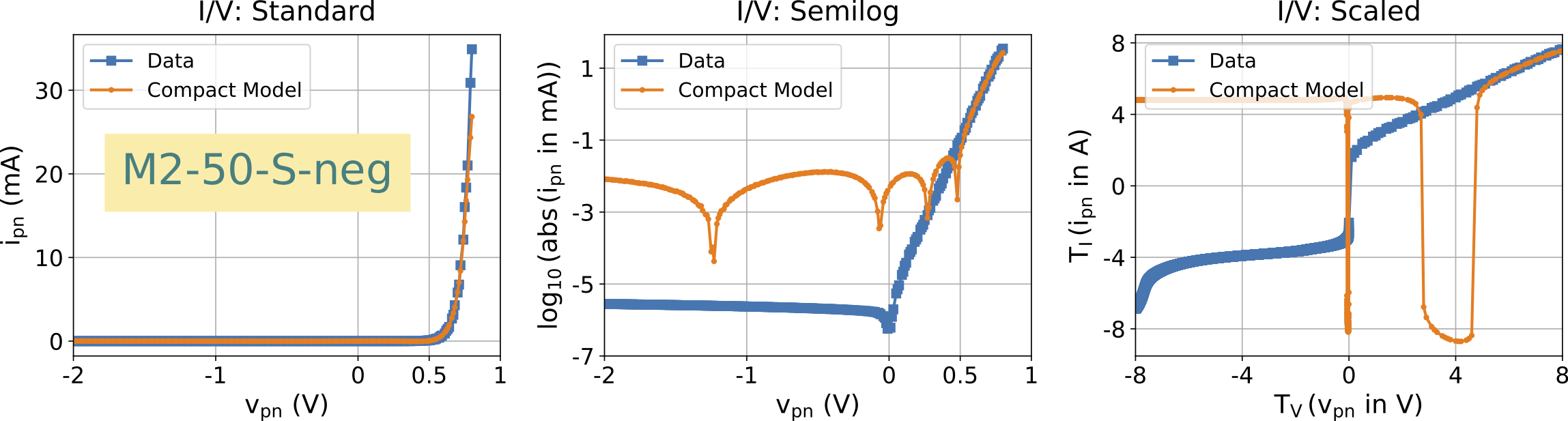}
\caption{\small Typical $I\!-\!V$ characteristics of DNN compact models learned without data transformations \emph{and without} the non-negative kernel constraint enforced: M1-50-S-neg (top) and M2-50-S-neg (bottom). These results highlight the importance of enforcing the non-negative kernel constraint and show that increasing the depth cannot compensate for the lack of this constraint. 
}
\label{fig:notneg-notx-noty}
\end{figure}

Figures~\ref{fig:notx-noty}--\ref{fig:notneg-notx-noty} illustrate these
failures using representative examples drawn from the 240 combinations examined
in this work. Figure~\ref{fig:notx-noty} shows typical results obtained with
eLU, sigmoid and tanh activation functions for networks having one hidden layer
and enforcing non-negative kernel constraints.  Figures
\ref{fig:E-notx-noty-1vs2} and \ref{fig:S-notx-noty-1vs2} compare the first two
examples from Figure \ref{fig:notx-noty} with networks having one additional
hidden layer. While the standard data views on these plots suggest regression
fit improvements as a result of introducing an extra hidden layer, the semilog
and scaled data views reveal that this is not necessarily the case. In
particular, we observe significant qualitative changes in the behavior of the
compact DNN models that are counterintuitive.

For example, the semilog view shows that M1-50-E does not have a zero switching
point whereas M1-50-S does. Interestingly, with two hidden layers, the
roles reverse and now M2-50-E exhibits a zero switching point, whereas M2-50-S
does not. Note, however, that the zero switching for M2-50-E occurs at a highly
non-physical value of almost $-2V$.

Figure \ref{fig:notneg-notx-noty} shows results with two DNN compact models that
do not enforce the non-negative kernel constraint.  In the case of M1-50-S-neg,
which has a single hidden layer, the unphysical nature of the model can already
be seen in the standard data view plot. The same view suggests that increasing
the number of hidden layers to two seems to dramatically improve the model.
However, the semilog and scaled data views again confirm that the resulting
model M2-50-S-neg is unphysical. In particular, we see that similar to
M1-50-S-neg, the model has multiple zero switching points, unlike the real
diode which has only one.

\begin{figure}[t!]
\centering
\includegraphics[width=0.85\textwidth]{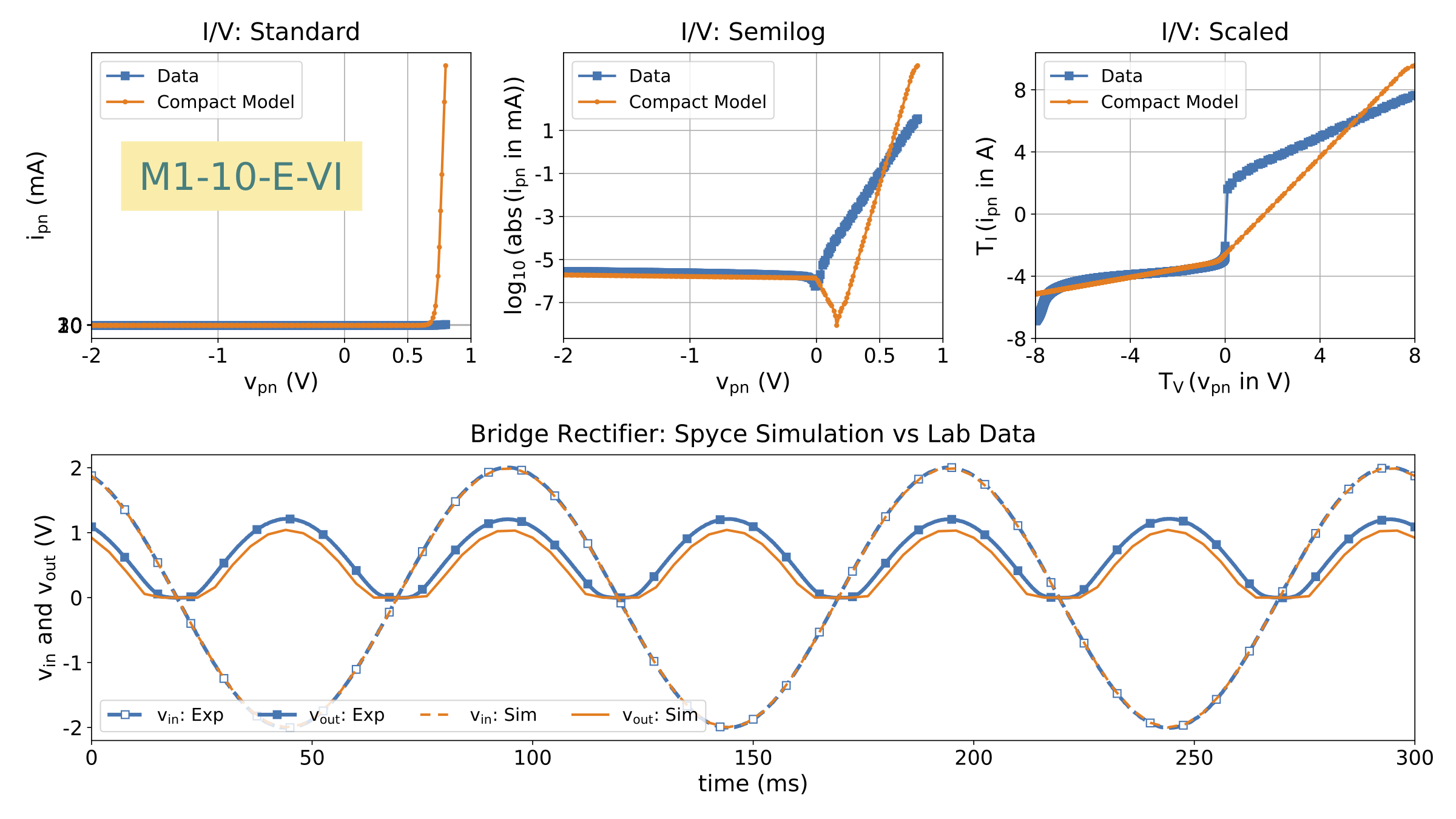}
\includegraphics[width=0.85\textwidth]{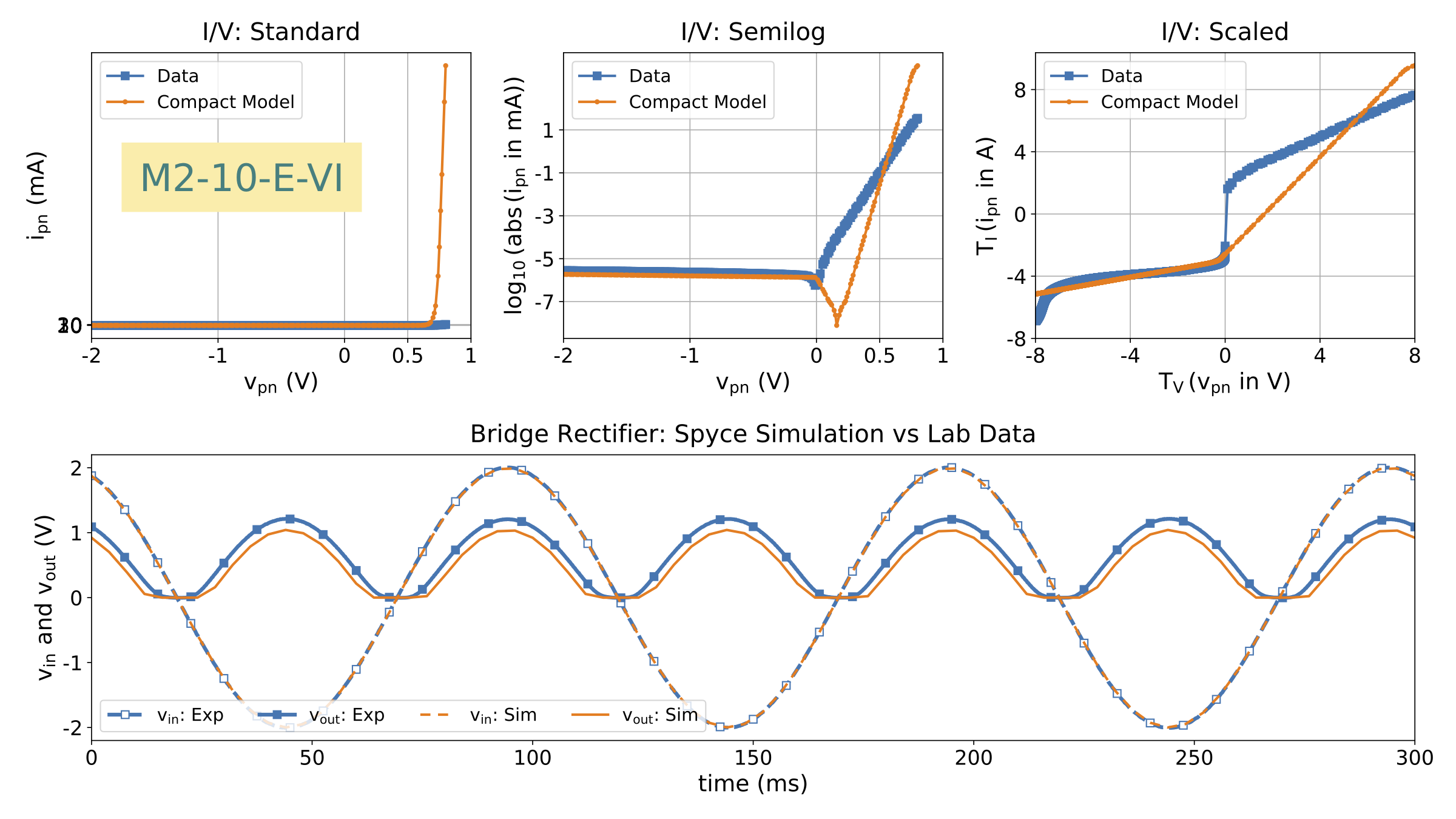}
\vspace{-1em}
\caption{\small Simulations comparing M1-10-E-VI (top) and M2-10-E-VI (bottom) compact DNN diode  models against laboratory data.}
\label{fig:M1-10-E-VI}
\end{figure}

\PBB{Our results indicate that} DNN performance markedly improves by switching
the training set to the transformed dataset $\mathcal{D}_{pn}^{VI}$.  However,
an important takeaway is that \PBB{not all  activation functions lead to
satisfactory compact models}. \PBB{In particular}, we found that eLU activation
performed much worse than sigmoid or tanh. For example, the top plot in Figure
\ref{fig:M1-10-E-VI} shows simulations of the M1-10-E-VI compact model; from the
standard $I\!-\!V$ characteristic plot, it is apparent that the model
greatly overestimates the current in the forward bias region. Indeed, at
$v_{pn}=0.8\textnormal{V}$, the current predicted by the model is so high that
the Y-axis tick labels on the standard view have all been compressed to the
point that they overlap one another. Still, the transformations of the training
set ensure that the model behaves well and has good convergence properties
during circuit simulation; the circuit simulation plot shows that the model even
gets reasonably close to matching laboratory data when it comes to circuit
behaviour. As in other cases, we did not observe a significant improvement in
the model when the number of layers was increased to two; see the results in the
bottom part of Figure \ref{fig:M1-10-E-VI}.

\begin{figure}[t!]
\centering
\includegraphics[width=0.85\textwidth]{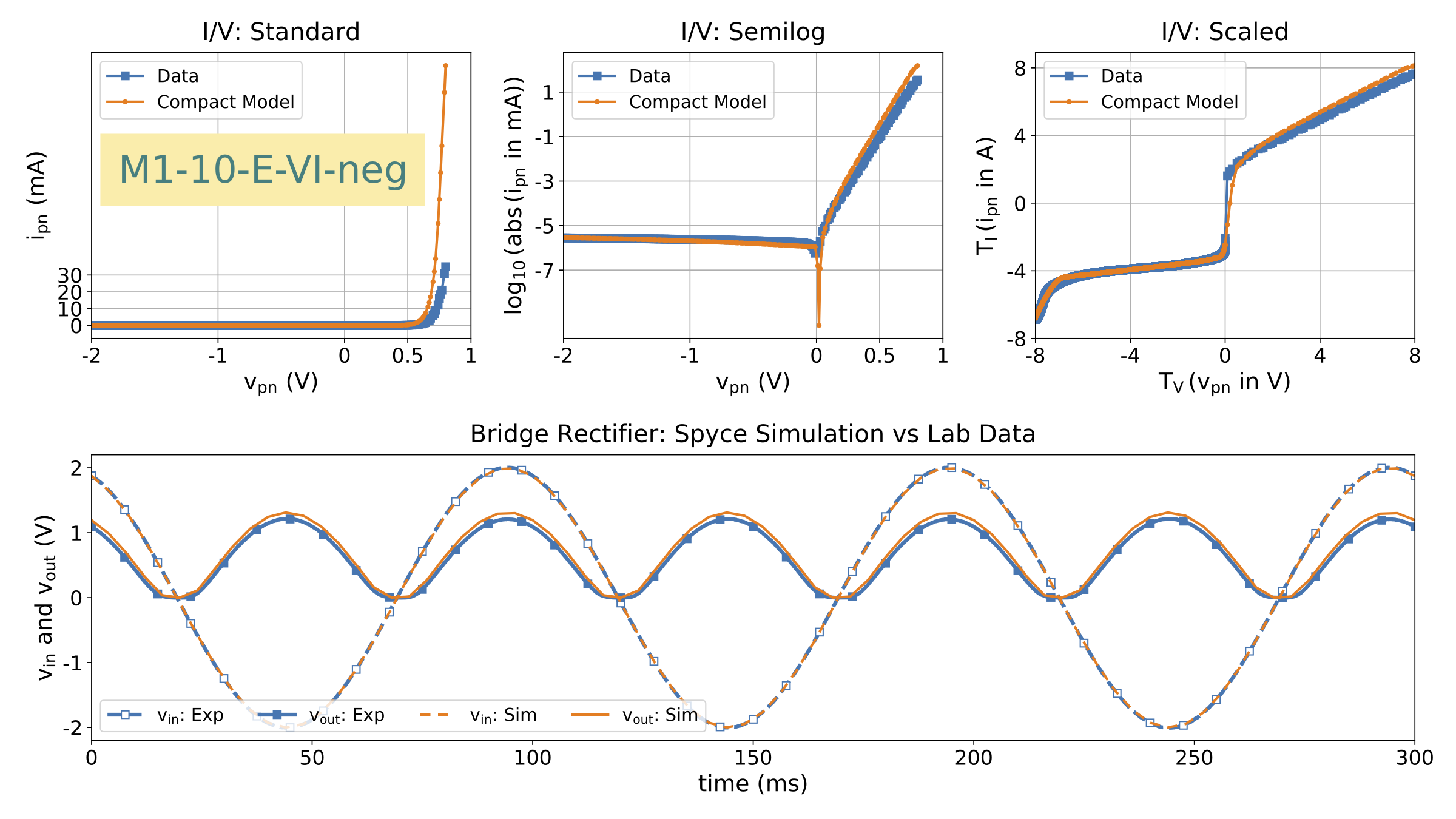}
\includegraphics[width=0.85\textwidth]{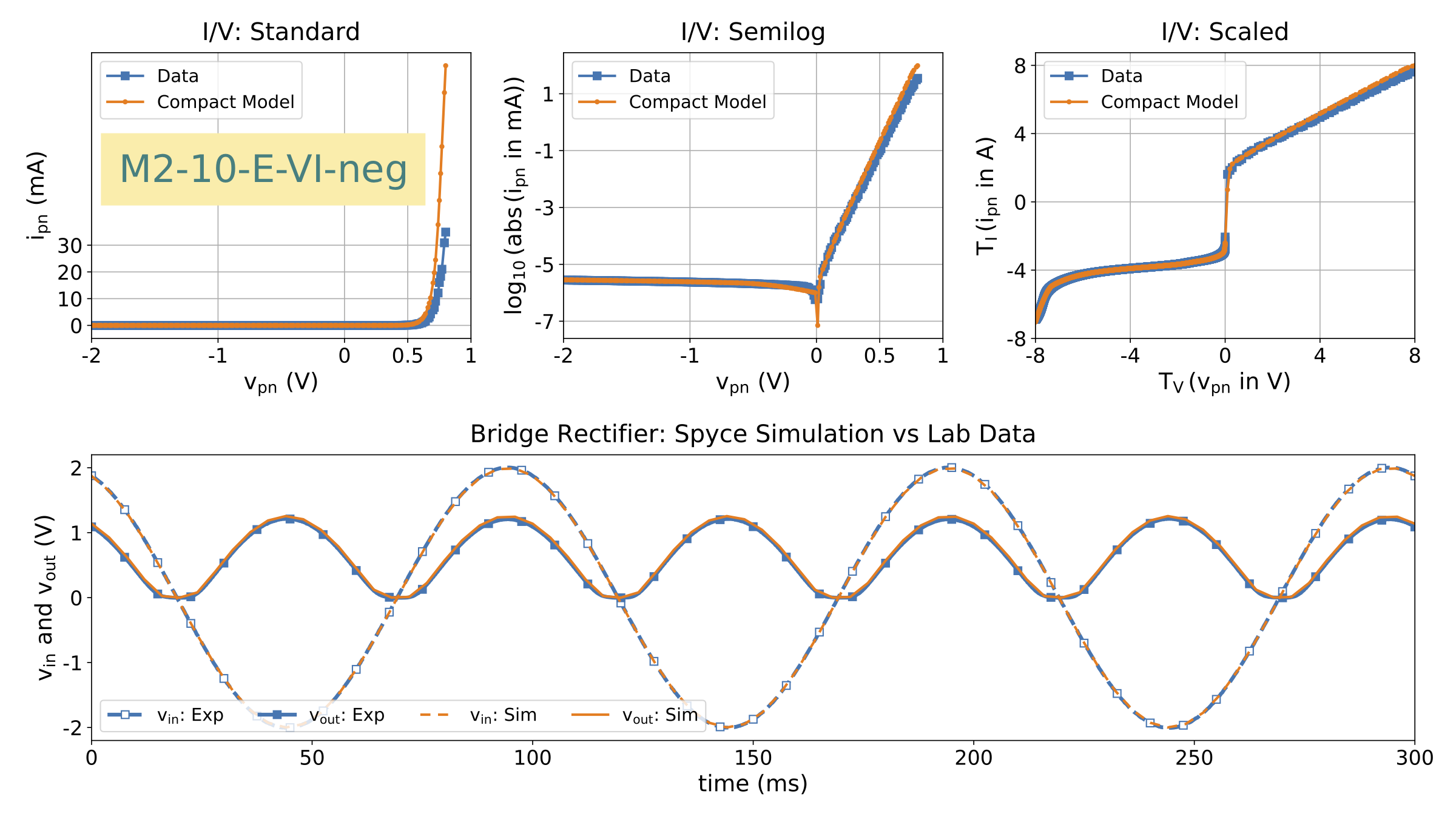}
\vspace{-1em}
\caption{\small Simulations comparing the M1-10-E-VI-neg (top) and M2-10-E-VI-neg (bottom) compact DNN diode model against laboratory data.}
\label{fig:M1-10-E-VI-neg}
\end{figure}

Figure \ref{fig:M1-10-E-VI-neg} shows an interesting corner case: with eLU
activation and data transformations, it looks like removing the non-negative
weight constraint actually helps the model become more accurate, both in its
$I\!\!-\!\!V$ characteristic and in circuit simulations. 

We hypothesize that this behavior is caused by eLU. Recall that eLU is the
identity function for positive inputs and an exponential converging to a
negative value for negative inputs. Thus, if the weights in the DNN are
constrained to be positive, the DNN's behavior is constrained to be close to
linear, which is a bad fit for a diode. Thus, in this special case, allowing
negative weights actually helps the DNN become more non-linear, and hence mimic
the data more closely.

With regard to eLU, it is worth pointing out that besides getting better models
without the non-negative kernel constraint, this was one of the few cases where
increasing the number of layers actually improved the model fit. As
Figure~\ref{fig:M1-10-E-VI-neg} shows, the M2-10-E-VI-neg model still
overestimates the current in the forward bias region, but to an extent
noticeably smaller than M1-10-E-VI-neg. This improvement in the $I\!\!-\!\!V$
characteristic of M2-10-E-VI-neg translates into more accurate circuit
simulations as well; in fact, the difference in the circuit simulations between
M2-10-E-VI-neg and M1-10-E-VI-neg is visible to the naked eye, as seen from the
two circuit simulations shown in Figure~\ref{fig:M1-10-E-VI-neg}. 

\begin{figure}[t!]
\centering
\includegraphics[width=0.85\textwidth]{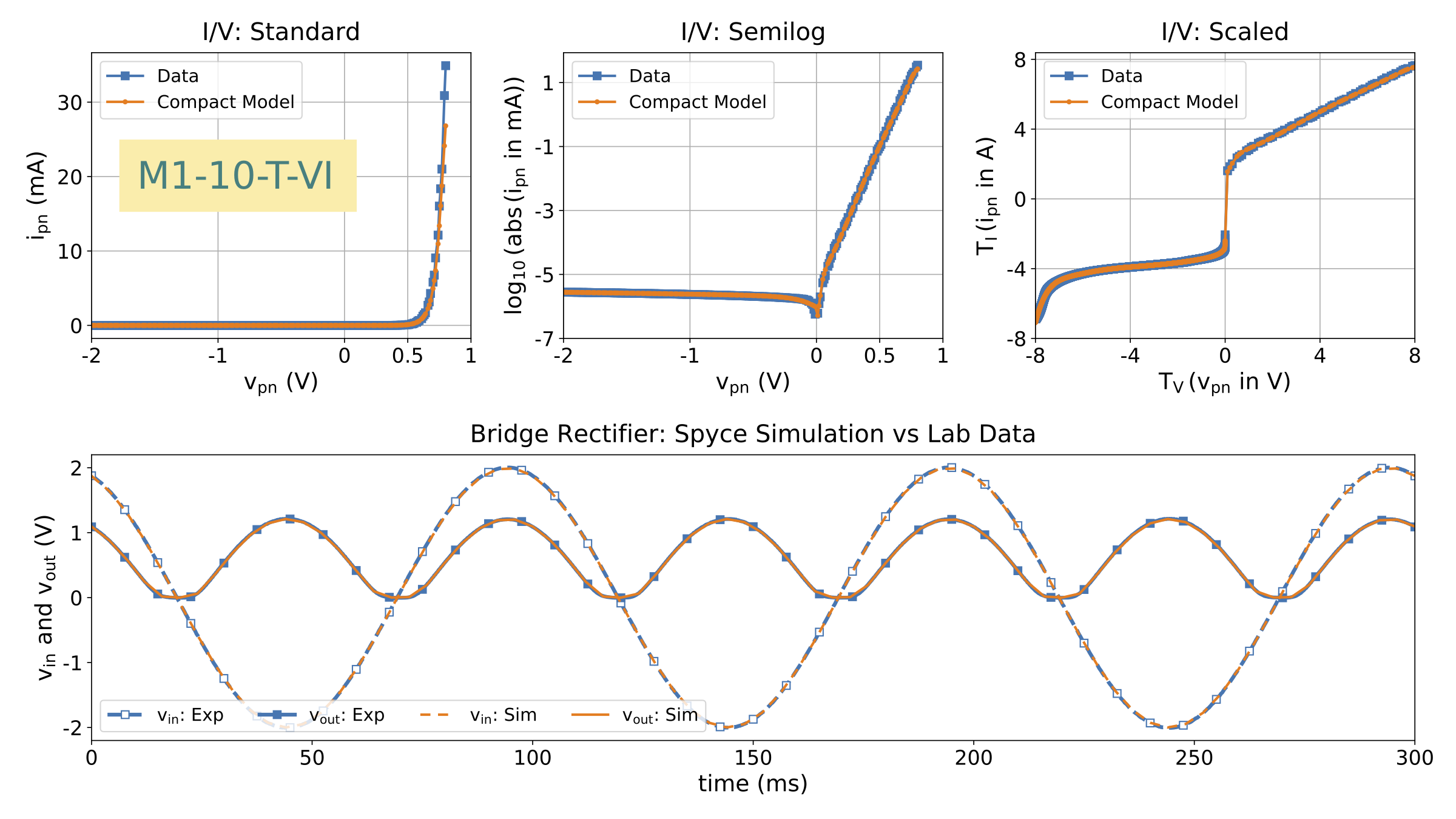}
\includegraphics[width=0.85\textwidth]{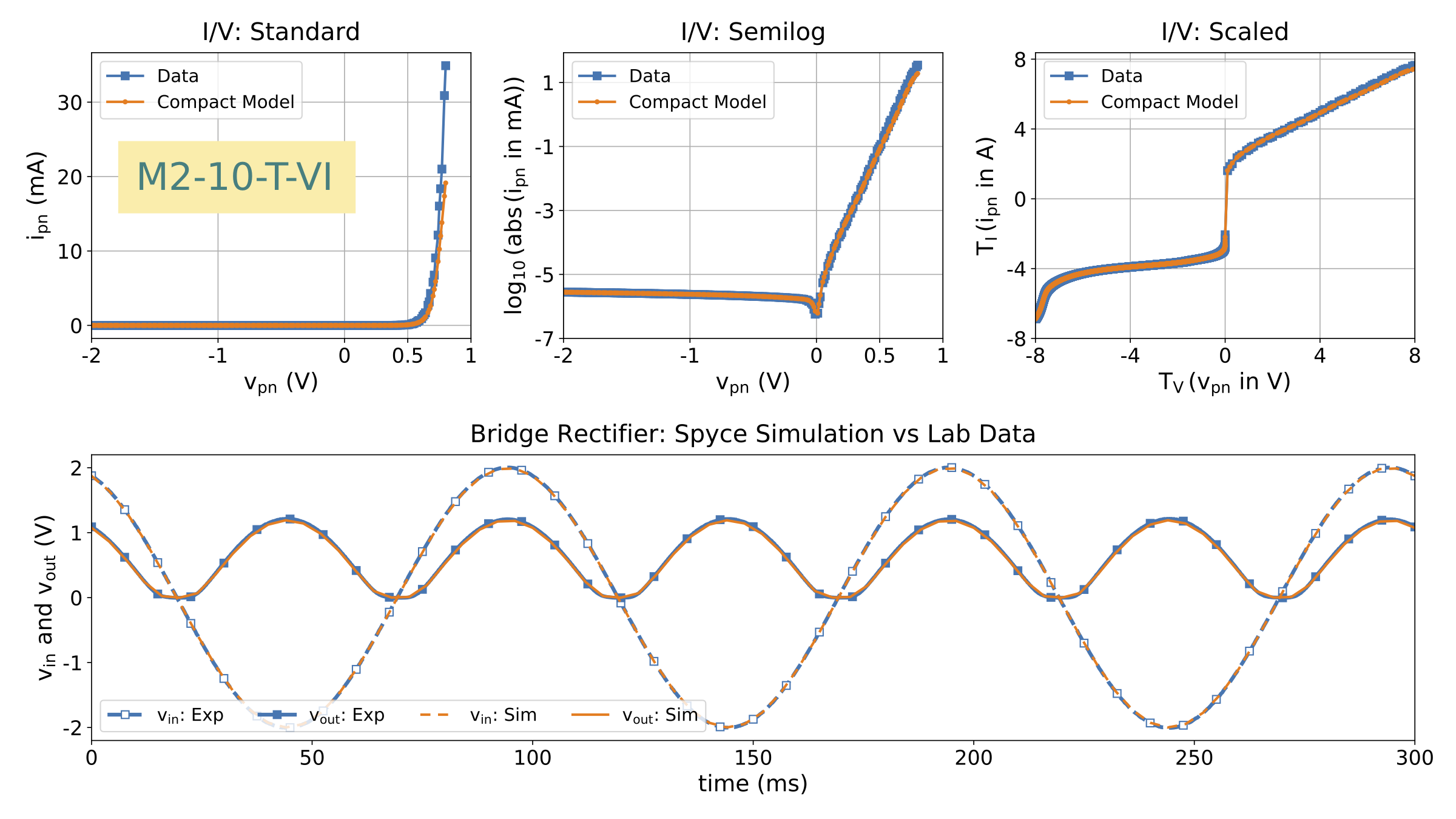}
\vspace{-1em}
\caption{\small Simulations comparing the M1-10-T-VI (top) and M2-10-T-VI (bottom) compact DNN diode models against laboratory data}
\label{fig:M1-10-T-VI}
\end{figure}

\begin{figure}[t!]
\centering
\includegraphics[width=0.85\textwidth]{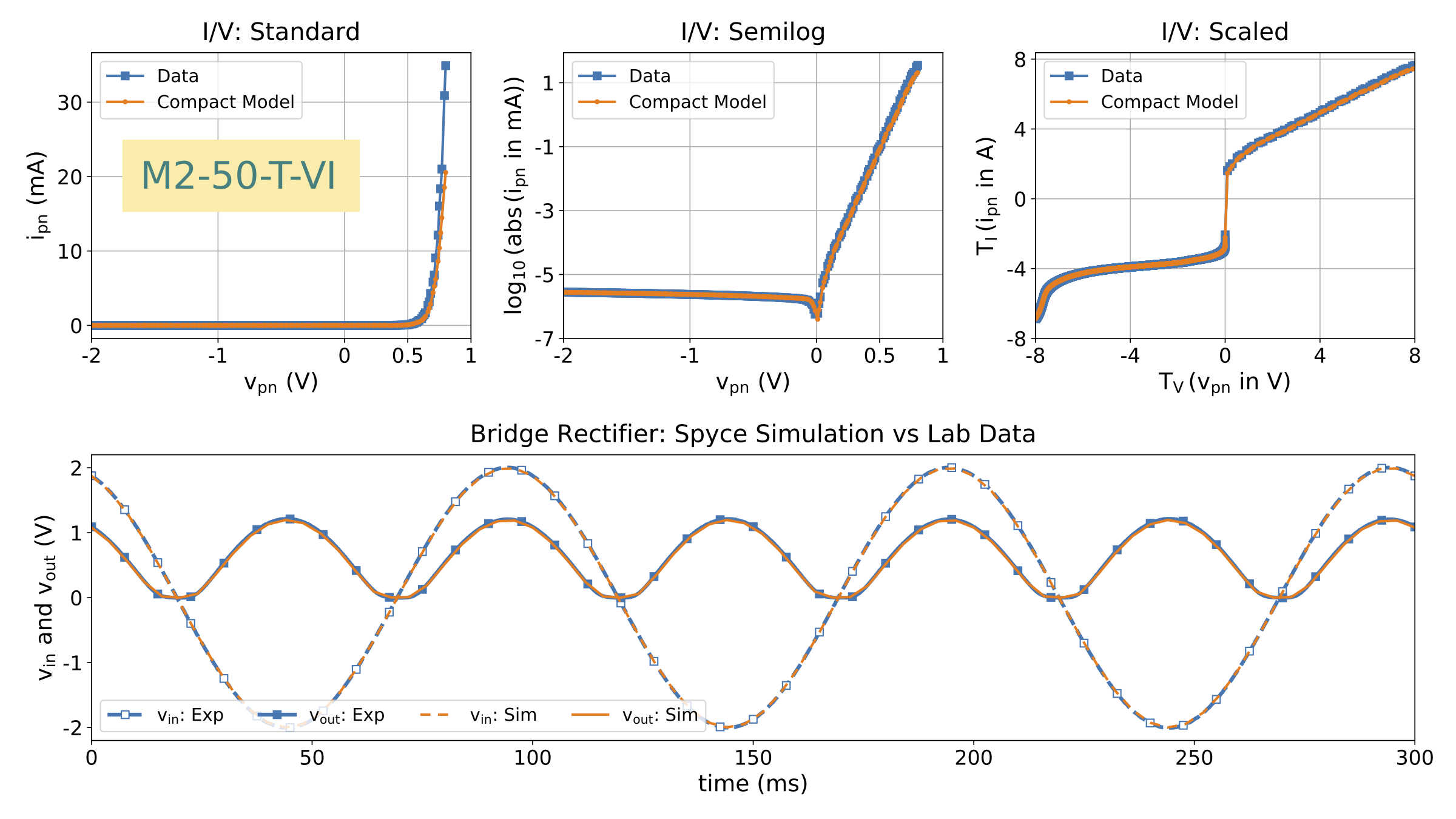}
\includegraphics[width=0.85\textwidth]{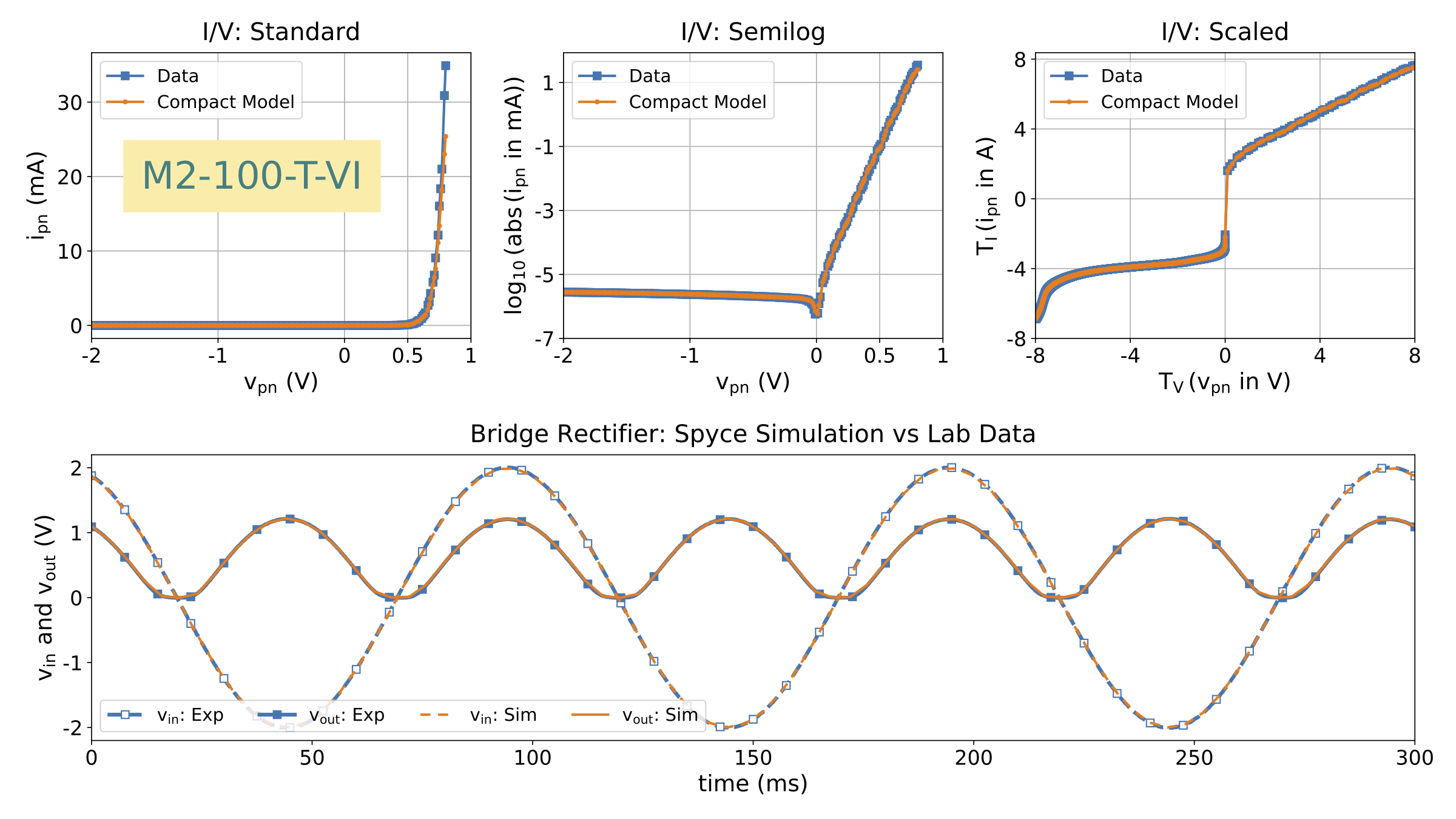}
\vspace{-1em}
\caption{\small Simulations comparing M2-50-T-VI (top) and M2-100-T-VI (bottom) compact DNN diode models against laboratory data}
\label{fig:M2-50--100-T-VI}
\end{figure}

By far, our most positive DNN finding is that combining sigmoid or tanh
activation with data transformations can produce extremely accurate,
\PBB{physically consistent}, and efficient DNN models. The top plot in Figure
\ref{fig:M1-10-T-VI} shows this for the M1-10-T-VI case; with just a single
hidden layer and 10 neurons, this tanh activated DNN is able to model all
aspects of the diode's $I\!\!-\!\!V$ characteristic very well, and perform very
accurately in circuit simulations as well. In fact, as evidenced by the results
in the bottom part of Figure~\ref{fig:M1-10-T-VI} and the plots in
Figure~\ref{fig:M2-50--100-T-VI}, increasing the number of layers and/or neurons
does not significantly increase model accuracy; the base model with just $10$
neurons and a single hidden layer is already so good that it is difficult to
improve upon it. We found that similar results held true with sigmoid
activation. Also, combining tanh/sigmoid with data transformations proved
very resilient to other training parameters; no matter what we chose for the
other parameters, the resulting DNN models were always accurate and converged
robustly (producing plots almost indistinguishable from those in Figures 
\ref{fig:M1-10-T-VI} and \ref{fig:M2-50--100-T-VI}).

Finally, we found that applying the data transformation to $i_{pn}$ was far more
critical than applying it to $v_{pn}$.  \PBB{In fact, the performance of the DNN
compact model showed only minor degradation (often invisible in plots) when we
applied $T_I$ but omitted applying $T_V$ to the training dataset.}

To summarize, the examples in
Figures~\ref{fig:notx-noty}--\ref{fig:notneg-notx-noty} and Figures
\ref{fig:M1-10-E-VI}--\ref{fig:M2-50--100-T-VI} highlight the following key
points:

\begin{itemize}
\item{Data transformations are critical to obtaining \PBB{accurate and 
physically consistent} DNN compact models;}
\item{The positive weight constraint enforcing monotonicity of $i_{pn}$
with respect to $v_{pn}$ is crucial; without this, the resulting DNN is often 
highly unphysical. One exception is eLU activation, which produces better 
models without this constraint; see Figures \ref{fig:M1-10-E-VI} and 
\ref{fig:M1-10-E-VI-neg}.}
\item{Generally, increasing the number of layers and/or neurons is not 
guaranteed to improve the model and can prompt significant qualitative changes 
in its behavior.}
\end{itemize}

Also, we note that even if the standard $I\!\!-\!\!V$ characteristic of a
data-driven compact model appears to match laboratory data satisfactorily, the
characteristic needs to be viewed on a semilog plot, as well as on a $(T_{V},
T_{I})$ scaled plot, to assess how well the model represents the physics of the
device. For example, although all $3$ models in Figure~\ref{fig:notx-noty} look
reasonable on the standard plot, the semilog and scaled plots reveal unphysical
behaviours such as current overestimation in both forward and reverse bias, as
well as the absence of or shifts in the zero crossing point.

\section{Conclusions}\label{sec:conclude}

In this paper, we investigated three different regression approaches to develop
data-driven compact models for a 1N4148 diode from laboratory measurements. The
first two (TBI via cubic splines and GMLS) are examples of local parametric and
non-parametric regression models.  Simulation results demonstrate that both of
these approaches deliver accurate and physically consistent compact device
models that show excellent agreement with laboratory $I\!-\!V$ measurements in
all device operational regimes. Furthermore, both compact models performed
robustly in circuit simulations; the simulated output of a bridge rectifier
circuit was in excellent agreement with laboratory measurements.  The TBI model
has higher memory requirements than GMLS but faster model evaluation. The use of
the performant Compadre toolkit however enables highly efficient GMLS
computation. As a result, both models can be deemed appropriate for the
one-junction device considered in this work.

For two-junction devices that will be the subject of a forthcoming paper, TBI
via multivariate splines will require data on a rectangular grid, whereas GMLS
will not have such a restriction. For such devices, GMLS may be more appropriate
for scattered electrical measurements.  

Our experiences with DNNs highlight the potential of this regression technique
for compact model development. In particular, using data transformations and
sigmoid/tanh activation, we were able to accurately regress a complex dataset
spanning multiple scales and comprising more than 9000 data points by a shallow
network with just 10 neurons. This makes our DNN compact model by far the most
memory efficient of all three kinds of data-driven models considered here;
unlike DNNs, TBI and GMLS models both require memory of the order of the size of
the entire dataset. This could be a significant advantage for DNNs when modeling
devices with more than one p-n junction. DNN models are also computationally
efficient: being a global regression, DNNs do not require dataset searches as
TBI and GMLS models do; their main cost is a few evaluations of their non-linear
activation functions.

At the same time, applying DNNs to approximate physics-based models requires
deeper understanding of their properties as regression tools. Although DNNs
potentially have the best generalizability of all the models considered in this
work, their regression accuracy depends on a complex interplay between depth,
width, activation functions, loss functions, constraint operators,~\eetc{}.
Although we have gained some insights on training DNNs for compact model
development (as described in Section~\ref{sec:results}), we believe that our
understanding is far from complete, and that devices with more than two
terminals will pose significant additional challenges \PBB{in identifying the
best combinations of} data transformations, architectures, constraint
operators,~\eetc{}.

Furthermore, circuit simulations underscore the importance of ensuring correct
physical behavior from data-driven compact models. For example, a good fit in the
forward bias region is not enough to ensure robust and physically correct
circuit simulations; this requires accurate representation of the zero crossing
and the leakage current in the reverse bias regime as well. Without these
features, circuit simulations are likely to be unphysical at best, and 
completely meaningless at worst. 

The results in this paper suggest that \emph{data transformations} are currently
the most effective heuristics for achieving acceptable accuracy in data-driven
DNN compact diode models. The transformations developed in this work aim to
reduce the vast difference in scales present in the $I\!-\!V$ diode
characteristic curve, which enables good data fits using a standard MSE loss
function. A custom loss function that adapts to multiscale data is another
potential option to improve DNN fit. We plan to pursue this work  in the future. 


\section*{Acknowledgments}{\small
Supported by the Laboratory Directed Research and Development program at Sandia
National Laboratories, a multimission laboratory managed and operated by
National Technology and Engineering Solutions of Sandia, LLC., a wholly owned
subsidiary of Honeywell International, Inc., for the U.S. Department of Energy's
National Nuclear Security Administration under contract DE-NA-0003525. This
paper describes objective technical results and analysis. Any subjective views
or opinions that might be expressed in the paper do not necessarily represent
the views of the U.S. Department of Energy or the United States Government.

The work of P.~Bochev has  also been supported by  the U.S. Department of
Energy, the Office of Science,  and the Office of Advanced Scientific Computing
Research under Award Number DE-SC-0000230927, as well as the Collaboratory on
Mathematics and Physics-Informed Learning Machines for Multiscale and
Multiphysics Problems (PhILMs) project.
}


\begin{thebibliography}{99}

\bibitem{Nagel_75_THESIS}
L.~W. Nagel.
\newblock {\em {SPICE2: A computer program to simulate semiconductor
  circuits}}.
\newblock PhD thesis, EECS Department, University of California, Berkeley,
  1975.

\bibitem{Keiter_19_TECHREPORT}
E.~R. Keiter, K.~V. Aadithya, T.~Mei, T.~V. Russo, R.~L. Schiek, P.~E.
  Sholander, H.~K. Thornquist, and J.~C. Verley.
\newblock {Xyce parallel electronic simulator: Users' guide, version 6.11}.
\newblock Technical Report SAND2019-5949, Sandia National Laboratories,
  Albuquerque, NM, 2019.

\bibitem{Keiter_19a_TECHREPORT}
E.~R. Keiter, K.~V. Aadithya, T.~Mei, T.~V. Russo, R.~L. Schiek, P.~E.
  Sholander, H.~K. Thornquist, and J.~C. Verley.
\newblock {Xyce parallel electronic simulator: Reference guide, version 6.11}.
\newblock Technical Report SAND2019-5950, Sandia National Laboratories,
  Albuquerque, NM, 2019.

\bibitem{charon}
\url{https://charon.sandia.gov/index.html}.

\bibitem{BSIM1}
Y.~S. Chauhan, D.~D. Lu, S.~Venugopalan, S.~Khandelwal, J.~P. Duarte,
  N.~Payvadosi, A.~Niknejad, and C.~Hu.
\newblock {\em {FinFET modeling for IC simulation and design: Using the
  BSIM-CMG standard}}.
\newblock Academic Press, 2015.

\bibitem{BSIM2}
W.~Liu and C.~Hu.
\newblock {\em {BSIM4 and MOSFET modeling for IC simulation}}.
\newblock World Scientific, 2011.

\bibitem{BSIM3}
W.~Liu and C.~Hu.
\newblock {BSIM3v3 MOSFET model}.
\newblock 9(03):671--701, 1998.

\bibitem{1N4148_1}
\url{https://en.wikipedia.org/wiki/1N4148_signal_diode}.

\bibitem{1N4148_2}
\url{https://www.diodes.com/assets/Datasheets/ds12019.pdf}.

\bibitem{deBoor}
C.~de~Boor.
\newblock {\em {A Practical Guide to Splines}}.
\newblock Springer-Verlag New York, 1978.

\bibitem{Gupta_17_INPROC}
A.~Gupta, T.~Wang, A.~G. Mahmutoglu, and J.~Roychowdhury.
\newblock {STEAM: Spline-based tables for efficient and accurate device
  modelling}.
\newblock In {\em ASPDAC '17: The $22^{nd}$ Asia and South Pacific Design
  Automation Conference}, pages 463--468, 2017.

\bibitem{Gupta_18_THESIS}
A.~Gupta.
\newblock {Table-based device modeling: Methods and applications}.
\newblock Master's thesis, EECS Department, University of California, Berkeley,
  2018.

\bibitem{Wendland_04_BOOK}
H.~Wendland.
\newblock {\em {Scattered data approximation}}.
\newblock Cambridge University Press, 2004.

\bibitem{Chen_17_JEM}
C.~Jiun-Shyan, H.~Michael, and C.~Sheng-Wei.
\newblock {Meshfree methods: Progress made after 20 years}.
\newblock 143(4):04017001, 2017.

\bibitem{Slattery_16_JCP}
S.~R. Slattery.
\newblock {Mesh-free data transfer algorithms for partitioned multiphysics
  problems: Conservation, accuracy, and parallelism}.
\newblock {\em Journal of Computational Physics}, 307:164--188, 2016.

\bibitem{Bungartz_16_CF}
H.-J. Bungartz, F.~Lindner, B.~Gatzhammer, M.~Mehl, K.~Scheufele, A.~Shukaev,
  and B.~Uekermann.
\newblock {PRECICE: A fully parallel library for multi-physics surface
  coupling}.
\newblock {\em Computers \& Fluids}, 141:250--258, 2016.
\newblock Advances in Fluid-Structure Interaction.

\bibitem{Goodfellow_16_BOOK}
I.~Goodfellow, J.~Bengio, and A.~Courville.
\newblock {\em {Deep Learning}}.
\newblock The MIT Press, 2016.

\bibitem{LeCun_15_Nature}
Y.~LeCun, Y.~Bengio, and G.~Hinton.
\newblock {Deep learning}.
\newblock {\em Nature}, 521:436--444, 2015.

\bibitem{Bar-Sinai_19_PNAS}
Y.~Bar-Sinai, S.~Hoyer, J.~Hickey, and M.~P. Brenner.
\newblock {Learning data-driven discretizations for partial differential
  equations}.
\newblock {\em Proceedings of the National Academy of Sciences},
  116(31):15344--15349, 2019.

\bibitem{Raissi_17_ArXiv}
M.~Raissi, P.~Perdikaris, and G.~E. Karniadakis.
\newblock {Physics informed deep learning (part 1): Data-driven solutions of
  non-linear partial differential equations}.
\newblock {\em arXiv preprint arXiv:1711.10561}, 2017.

\bibitem{Zaabab_94_IEEE_MTT-S}
A.~H. Zaabab, Q.-J. Zhang, and M.~S. Nakhla.
\newblock {Analysis and optimization of microwave circuits and devices using
  neural network models}.
\newblock In {\em IEEE MTT-S International Microwave Symposium Digest},
  volume~1, pages 393--396, May 1994.

\bibitem{Zaabab_95_IEEE_TMTT}
A.~H. Zaabab, Q.-J. Zhang, and M.~S. Nakhla.
\newblock {A neural network modeling approach to circuit optimization and
  statistical design}.
\newblock {\em IEEE Transactions on Microwave Theory and Techniques},
  43(6):1349--1358, June 1995.

\bibitem{Meijer_96_THESIS}
P.~B.~L. Meijer.
\newblock {\em {Neural network applications in device and sub-circuit modelling
  for circuit simulation}}.
\newblock PhD thesis, Department of Chemical Engineering and Chemistry,
  Technische Universiteit Eindhoven, 1996.

\bibitem{Litovski_97_SPT}
V.~B. Litovski, {\v Z}.~Mr{\v c}arica, and T.~Ili{\'c}.
\newblock {Simulation of non-linear magnetic circuits modelled using artificial
  neural network}.
\newblock {\em Simulation Practice and Theory}, 5(6):553--570, 1997.

\bibitem{Andrejevic_03_JAC}
M.~Andrejevi{\'c} and V.~B. Litovski.
\newblock {Electronic circuits modeling using artificial neural networks}.
\newblock {\em Journal of Automatic Control, University of Belgrade},
  13(1):31--37, 2003.

\bibitem{Chen_06_INPROC}
X.~Chen, G.~F. Wang, W.~Zhou, Q.~L. Zhang, and J.~F. Xu.
\newblock {Application of neural networks for integrated circuit modeling}.
\newblock In {\em Advances in Neural Networks}, volume 3973, pages 1304--1312,
  2006.

\bibitem{Gorissen_09_NCA}
D.~Gorissen, L.~D. Tommasi, and K.~Crombecq.
\newblock {Sequential modeling of a low noise amplifier with neural networks
  and active learning}.
\newblock {\em Neural Computing \& Applications}, 18:485--494, 2009.

\bibitem{Chen_17_INPROC}
Z.~Chen, M.~Raginsky, and E.~Rosenbaum.
\newblock {Verilog-A compatible recurrent neural network model for transient
  circuit simulation}.
\newblock In {\em EPEPS '17: The $26^{th}$ IEEE Conference on Electrical
  Performance of Electronic Packaging and Systems}, pages 1--3, 2017.

\bibitem{Zaabab_97_IEEE_TMTT}
A.~H. Zaabab, Q.-J. Zhang, and M.~S. Nakhla.
\newblock {Device and circuit-level modeling using neural networks with faster
  training based on network sparsity}.
\newblock {\em IEEE Transactions on Microwave Theory and Techniques},
  45(10):1696--1704, Oct 1997.

\bibitem{Hammouda_08_AJAS}
H.~B. Hammouda, M.~Mhiri, Z.~Gafsi, and K.~Besbes.
\newblock {Neural-based models of semiconductor devices for SPICE simulator}.
\newblock {\em American Journal of Applied Sciences}, pages 385--391, 2008.

\bibitem{Lei_18_INPROC}
Y.~Lei, X.~Huo, and B.~Yan.
\newblock {Deep neural network for device modeling}.
\newblock In {\em EDTM '18: The $2^{nd}$ IEEE Electron Devices Technology and
  Manufacturing Conference}, pages 154--156, 2018.

\bibitem{Li_16_IEEE_JESSCDC}
M.~Li, O.~{\.I}rsoy, C.~Cardie, and H.~G. Xing.
\newblock {Physics-inspired neural networks for efficient device compact
  modeling}.
\newblock {\em IEEE Journal on Exploratory Solid-State Computational Devices
  and Circuits}, 2:44--49, Dec 2016.

\bibitem{Kuberry_19_MISC}
P.~Kuberry, P.~Bosler, and N.~Trask.
\newblock {Compadre toolkit}, February 2019.

\bibitem{Tensorflow_15_MISC}
M.~Abadi, A.~Agarwal, P.~Barham, E.~Brevdo, Z.~Chen, C.~Citro, G.~S. Corrado,
  A.~Davis, J.~Dean, M.~Devin, S.~Ghemawat, I.~Goodfellow, A.~Harp, G.~Irving,
  M.~Isard, Y.~Jia, R.~Jozefowicz, L.~Kaiser, K.~Kudlur, J.~Levenberg,
  D.~Man\'{e}, R.~Monga, S.~Moore, D.~Murray, C.~Olah, M.~Schuster, J.~Shlens,
  B.~Steiner, I.~Sutskever, K.~Talwar, P.~Tucker, V.~Vanhoucke, V.~Vasudevan,
  F.~Vi\'{e}gas, O.~Vinyals, W.~Warden, M.~Wattenberg, M.~Wicke, Y.~Yu, and
  X.~Zheng.
\newblock {TensorFlow: Large-scale Machine Learning on heterogeneous systems},
  2015.
\newblock Software available from tensorflow.org.

\bibitem{Wang_16_IEEE_MTT-S}
T.~Wang and J.~Roychowdhury.
\newblock {Multiphysics modelling and simulation in Berkeley MAPP}.
\newblock In {\em NEMO '16: The IEEE MTT-S International Conference on
  Numerical Electromagnetic and Multiphysics Modeling and Optimization}, pages
  1--3, July 2016.

\bibitem{Griewank2008EDP}
A.~Griewank and A.~Walther.
\newblock {\em {Evaluating derivatives: Principles and techniques of
  algorithmic differentiation}}.
\newblock SIAM, 2 edition, 2008.

\bibitem{shockley1949theory}
W.~Shockley.
\newblock {The theory of p-n junctions in semiconductors and p-n junction
  transistors}.
\newblock {\em Bell System Technical Journal}, 28(3):435--489, 1949.

\bibitem{trefethen2013approximation}
L.~N. Trefethen.
\newblock {\em {Approximation theory and approximation practice}}, volume 128.
\newblock SIAM, 2013.

\bibitem{Mirzaei_12_IMAJNA}
D.~Mirzaei, R.~Schaback, and M.~Dehghan.
\newblock {On generalized moving least squares and diffuse derivatives}.
\newblock {\em IMA Journal of Numerical Analysis}, 32(3):983--1000, 2012.

\bibitem{Golub_96_BOOK}
G.~Golub and C.~F.~V. Loan.
\newblock {\em {Matrix computations}}.
\newblock Johns Hopkins University Press, 3 edition, 1996.

\bibitem{Opschoor_19_TECHREPORT}
J.~A.~A. Opschoor, P.~C. Petersen, and C.~Schwab.
\newblock {Deep ReLU networks and high-order finite element methods}.
\newblock Research Report 2019-07, ETH Zurich, January 2019.

\end{thebibliography}
\end{document}